\pdfoutput=1
\PassOptionsToPackage{table}{xcolor}
\documentclass[11pt]{article}

\usepackage[final]{acl}

\usepackage{times}
\usepackage{latexsym}

\usepackage{color}
\usepackage{bm}
\usepackage{mathtools}
\usepackage{amsfonts} 
\usepackage{amsthm}
\usepackage{array,multirow}
\usepackage{tabularx}
\usepackage{arydshln}
\usepackage{adjustbox}
\usepackage{balance}
\usepackage{wrapfig}
\usepackage{xcolor, pifont}
\usepackage{soul}
\usepackage{booktabs}
\usepackage{kotex}
\hypersetup{colorlinks} 

\definecolor{aliceblue}{rgb}{0.94, 0.97, 1.0}
\definecolor{lightgray}{rgb}{0.86, 0.86, 0.86}
\definecolor{blanchedalmond}{rgb}{1.0, 0.92, 0.8}
\definecolor{forestgreen}{rgb}{0.13,0.55,0.13}
\newcommand{\cmark}{\ding{51}}%
\newcommand{\eg}{\textit{e}.\textit{g}.}
\newcommand{\ie}{\textit{i}.\textit{e}.}
\newcommand{\red}[1]{{\color{red}#1}}

\usepackage{listings}
\usepackage{xcolor}

\usepackage[T1]{fontenc}

\usepackage[utf8]{inputenc}

\usepackage{microtype}

\usepackage{inconsolata}

\usepackage{graphicx}

\title{R-VLM: Region-Aware Vision Language Model for Precise GUI Grounding}

\author{\textbf{Joonhyung Park}$^{\dagger}$\thanks{This work was done when Joonhyung Park was an intern at Amazon. $^{**}$Peng Tang is the corresponding author.} \quad \textbf{Peng Tang}$^{\ddagger}$$^{**}$ \quad \textbf{Sagnik Das}$^{\ddagger}$ \quad \textbf{Srikar Appalaraju}$^{\ddagger}$ \\ \textbf{Kunwar Yashraj Singh}$^{\ddagger}$ \quad \textbf{R. Manmatha}$^{\ddagger}$ \quad \textbf{Shabnam Ghadar}$^{\ddagger}$\\
$^{\dagger}$KAIST \quad $^{\ddagger}$AWS AI Labs\\
\texttt{deepjoon@kaist.ac.kr}, \texttt{\{tangpeng723, manmatha\}@gmail.com} \\
\texttt{\{sagnikd, srikara, sinkunwa, shabnam\}@amazon.com}}

\begin{document}
\maketitle

\begin{abstract}
Visual agent models for automating human activities on Graphical User Interfaces (GUIs) have emerged as a promising research direction, driven by advances in large Vision Language Models (VLMs). A critical challenge in GUI automation is the precise grounding of interface elements across diverse platforms. Existing vision-only GUI agents directly ground elements from large and cluttered screenshots, requiring them to process substantial irrelevant information that compromises their accuracy. In addition, these approaches typically employ basic cross-entropy loss for learning grounding objectives, which fails to effectively capture grounding quality compared to established object detection metrics like Intersection-over-Union (IoU). To address these issues, we introduce R-VLM, a novel GUI grounding approach that leverages zoomed-in region proposals for precise element localization. We also propose an IoU-aware objective function that facilitates model convergence toward high IoU predictions. Our approach bridges the gap between VLMs and conventional object detection techniques, improving the state-of-the-art grounding accuracy by 13\% across diverse GUI platforms on the GUI grounding benchmarks ScreenSpot and AgentStudio. In addition, our R-VLM approach shows 3.2-9.7\% absolute accuracy improvements in GUI navigation tasks on the AITW and Mind2Web benchmarks.

\end{abstract}

\section{Introduction}
\label{sec:intro}
\begin{figure*}[ht]
    \centering
    \vspace{-0.1in}
    \captionsetup{type=figure, width=1.\linewidth}
    \includegraphics[width=1.\linewidth]{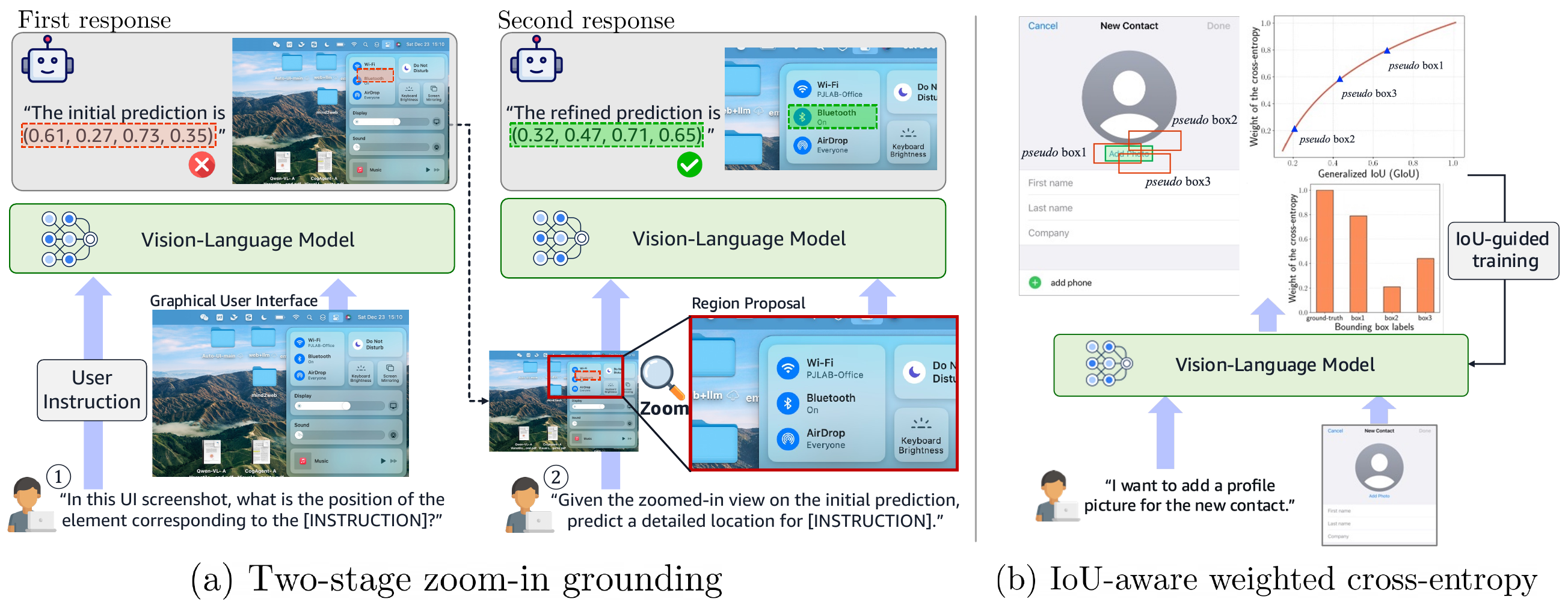}
    \vspace{-0.2in}
    \caption{Illustration of the Region-Aware Vision Language Model (R-VLM). Our approach consists of two modules for precise GUI grounding: (a) A two-stage zoom-in grounding process that refines predictions via a zoomed-in view of region proposal. After obtaining an initial prediction from the model using GUI screenshot and user instruction, which serves as a region proposal, we zoom-in around this region and pass it through the model again for second-stage grounding. (b) An IoU-aware weighted cross-entropy loss that provides a smooth learning signal based on the IoU value rather than strictly fitting to ground-truth bounding box. This loss assigns weights to pseudo bounding boxes according to their IoU value with ground-truth to emphasize high IoU grounding predictions.}
    \label{fig:main_figure}
    \vspace{-5mm}
\end{figure*}

Agent models operating on Graphical User Interface (GUI) have gained significant attention for their ability to automate cognitively demanding tasks on behalf of humans from high-level instructions \cite{gurreal,he2024webvoyager,cogagent,furutamultimodal}. Recent remarkable advancements in Vision Language Model's (VLM) ability to perceive visual elements and align language instructions to GUI objects have paved the way to automating tasks in real user interfaces.
However, most of the current approaches rely on hierarchical text modalities (\eg, HTML, accessibility tree, and DOM tree) to understand GUI layouts and interactable elements \cite{zheng2024gpt,gurreal,zhouwebarena,zhu2025turbocharging,Gao_2024_CVPR}.
This dependence on textual representation presents significant limitations for real-world applications, as it requires processing extensive contextual information and managing diverse GUI representations across different environments - web, operating systems, and desktop applications. Consequently, there is growing interest in developing vision-only agent models that operate solely on visual inputs (\ie, GUI screenshots).

Precise GUI grounding, \ie, accurately predicting the coordinates of the GUI elements,
is fundamental to the successful operation of visual agents. Complex GUI layouts often lead to inaccurate localization, resulting in failed action executions and compromised multi-step automation processes. This challenge motivates our central research question: \textit{How can VLMs be empowered to achieve accurate coordinate prediction of GUI elements given GUI screenshots and natural language instructions?}

Previous vision-only agents have introduced several techniques to enhance GUI grounding \cite{seeclick, cogagent,shaw2023pixels}. However, these approaches struggle to precisely locate GUI elements, mainly due to two fundamental limitations. First, these works attempt to predict grounding coordinates from entire screenshots, which typically contain complex layouts and diverse GUI elements spanning multiple scales. This inherent complexity in coordinate prediction, a challenge well-discussed in conventional object detection literature \cite{rcnn,fastrcnn,fasterrcnn}, substantially impedes the achievement of fine-grained grounding accuracy. Second, existing approaches employ basic cross-entropy loss for training the grounding model in the token space. While cross-entropy loss effectively optimizes specific token predictions by reducing the likelihood of all other tokens, it fails to provide meaningful learning signals that reflect the quality of grounding predictions, particularly when evaluated using standard object detection metrics such as Intersection-over-Union (IoU).

Focusing on the aforementioned limitations, we introduce the Region-aware Vision Language Model (\textbf{R-VLM}), a novel region-based approach that substantially improves GUI grounding accuracy by incorporating principles from established object detection algorithms. Our approach consists of two key components: a two-stage zoom-in grounding approach inspired by region proposal networks~\cite{fasterrcnn} (Figure~\ref{fig:main_figure}(a)), and an IoU-aware weighted cross-entropy loss function that emphasizes achieving high IoU grounding, similar to how conventional object detection models are optimized~\cite{rcnn,fastrcnn} (Figure~\ref{fig:main_figure}(b)).

In the two-stage zoom-in grounding approach, we first predict a bounding box that serves as a region proposal, followed by a zoom-in around this region to obtain a more fine-grained prediction. In addition, we also propose a simple yet effective recipe for zoom-in instruction tuning data generation for VLM fine-tuning, which further strengthens the two-stage zoom-in process.
For the IoU-aware weighted cross-entropy loss function, multiple pseudo Ground Truth (GT) bounding boxes are generated around the real GT.
The cross-entropy weights are dynamically adjusted during training according to the IoU values between the pseudo GT and real GT, providing continuous learning signals that guide the model toward higher IoU grounding predictions.
Furthermore, to mitigate computational overhead from additional pseudo GTs, we propose a cost-efficient way for the IoU-aware loss by modifying the attention map and relative positional encoding. This design maintains the effectiveness of our approach while reducing computational demands.

We apply our approach to the previous state-of-the-art approach, SeeClick~\cite{seeclick}.
Extensive experiments are conducted on the GUI grounding datasets ScreenSpot \cite{seeclick} and AgentStudio \cite{zheng2024agentstudio} as well as the GUI navigation datasets AITW \cite{rawles2024androidinthewild} and Mind2Web \cite{multimoda-mind2web}.
Our approach achieves 13\% absolute improvements in grounding accuracy across diverse GUI platforms (mobile, desktop, and web) on ScreenSpot and AgentStudio. Also, R-VLM outperforms the baseline by 3.2-9.7\% on the GUI navigation datasets AITW and Mind2Web. Notably, our approach can be seamlessly integrated with any VLM, and applying the two-stage zoom-in grounding alone to VLMs in a training-free manner yields substantial accuracy gains.

 
We summarize our contributions as follows.
\vspace{-0.2cm}
\begin{itemize}
\setlength\itemsep{0mm}
    \item We propose R-VLM, a novel region-based approach for precise GUI grounding that incorporates principles from conventional object detection algorithms into VLMs. R-VLM consists of two key components: a two-stage zoom-in grounding approach that leverages zoomed-in region proposals for fine-grained localization, and an IoU-aware weighted cross-entropy loss that emphasizes high IoU grounding predictions.
    \item We apply R-VLM to the previous state-of-the-art approach. Our approach obtains 13\% absolute GUI grounding accuracy improvements and 3.2-9.7\% absolute GUI navigation accuracy improvements compared to the previous state-of-the-art.
    \item We show that the two-stage zoom-in grounding approach can be applied to other VLMs in a training-free manner with substantial accuracy improvements.
\end{itemize}

\section{Related Work}
\label{sec:related_works}

\paragraph{Agents for GUI Automation}
With recent advancements in large language models, automating human activities on real-world Graphical User Interfaces (GUIs) has become increasingly feasible. Recent studies have pursued general-purpose autonomous GUI agents~\cite{mind2web,zhouwebarena,gao2023assistgui}. 
Although these efforts show promising results in automation tasks, they rely on structured documents, which necessitate processing verbose documents and limit dynamic operation across diverse platforms. To address these challenges, recent approaches propose agents that operate solely on visual inputs using vision-language models~\cite{shaw2023pixels,cogagent,seeclick,zhang2023you}. CogAgent~\cite{cogagent} introduces a dedicated pathway for high-resolution images, and SeeClick~\cite{seeclick} pretrains a model using a series of GUI grounding tasks. However, current vision-only agents train models with cross-entropy loss, treating numeric coordinates as discrete tokens, and do grounding directly from entire screenshots, leading to low-IoU grounding. We improve vision-language model grounding by enabling them to operate similarly to object detection algorithms with the IoU-aware training objective. 

\paragraph{Conventional Object Detection Approaches}
The region-based approach with IoU-aware loss is a well-established strategy in conventional object detection \cite{rcnn,fastrcnn,fasterrcnn,liu2016ssd}.
This approach obtains region proposals from models, 
such as Selective Search \cite{uijlings2013selective}, EdgeBoxes \cite{zitnick2014edge}, Region Proposal Network \cite{fasterrcnn} or sliding windows \cite{liu2016ssd} (a.k.a. anchor boxes).
Subsequently, the features of each region proposal are extracted by cropping the original image \cite{rcnn} or feature maps of the original image \cite{fastrcnn,fasterrcnn,liu2016ssd,patel2023simcon}.
The region proposal features are then fed into neural network layers with a classification head and a bounding box regression head to produce precise object detection results, where the bounding box regression head is trained with regression loss that is IoU-aware.
Our proposed R-VLM approach draws inspiration from conventional object detection approaches.
Specifically, R-VLM employs a two-stage zoom-in grounding mechanism, inspired by region proposal-based cropping, and incorporates an IoU-aware loss analogous to bounding box regression loss.
We adapt conventional object detection approaches to make them applicable to VLM-based GUI grounding. 
\section{Region-Aware Vision Language Model}
\label{sec:method}
We introduce the Region-aware Vision Language Model (R-VLM), a region-guided framework tailored for precise GUI localization with vision-language models (Figure~\ref{fig:main_figure}). First, we present analyses of existing visual agent models localizing GUI elements, uncovering challenges that need to be addressed (Section~\ref{subsec:prelim}). Building on these observations, we propose our method with two key components. (1) A two-stage zoom-in approach that derives fine-grained and accurate grounding results, even for small objects (\eg, icons), from the zoomed-in view of the GUI screenshot, enabling the model to function similarly to region proposal networks (Section~\ref{subsec:zoomingrounding}). (2) An Intersection-over-Union (IoU) guided objective function that provides the model with learning signals to comprehend the concept of IoU during training, where numeric values are typically treated as sequential language tokens (Section~\ref{subsec:iouloss}). 
Note that while we describe our approach with box prediction for GUI elements for brevity, it can be readily applied to point coordinate prediction tasks as well, see Appendix \ref{subsec:appx_pointpred}. 

\begin{figure}[t]
  \centering
  \includegraphics[width=\linewidth]{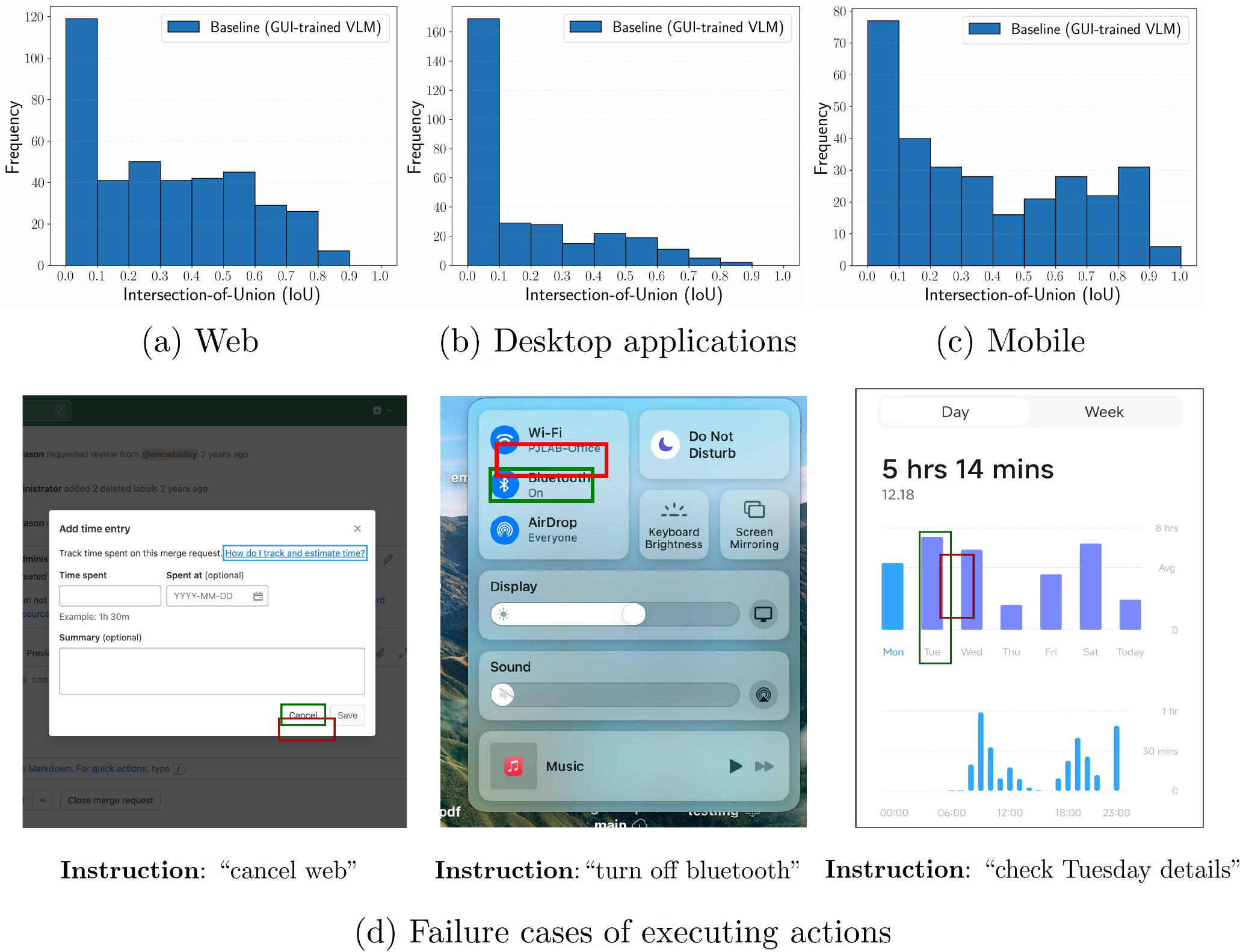}
  \caption{Histogram of IoU scores for predictions by the state-of-the-art approach SeeClick~\cite{seeclick} on the ScreenSpot benchmark show consistent low IoU across GUI environments: (a) Web, (b) Desktop applications, and (c) Mobile. (d) Qualitative examples of action execution failure cases, where the model captures nearby regions but fails to localize precisely.}
  \label{fig:problem_1}
  \vspace{-6mm}
\end{figure}

\subsection{Preliminary Study}
\label{subsec:prelim}

We briefly examine how existing visual agent model, a vision-language model pretrained with GUI grounding data, predicts GUI elements. First, we analyze the IoU histogram between predicted bounding boxes and corresponding ground truths (Figure~\ref{fig:problem_1} (a)-(c)) on ScreenSpot benchmark using a recent representative baseline~\cite{seeclick} that pretrains Qwen-VL~\cite{qwenvl} on a series of GUI grounding tasks. As shown in the figure, the model's predictions exhibit low-IoU patterns across GUI objects. These results align with failure cases commonly observed, where the center point of the predicted box does not fall within the ground-truth box (Figure~\ref{fig:problem_1} (d)). While the model captures nearby parts, it often fails to pinpoint exact bounding box locations (low IoU scores), suggesting limited grounding ability in vision-language models. Another type of failure involves mismatches between user instructions and GUI icons/widgets (\eg, ``Delete this mail" mapped to a trash can icon); however, we focus on the aforementioned issues here as the mismatch problem can be alleviated with curated GUI grounding data at scale.

\begin{figure}[t]
  \centering
  \includegraphics[width=\linewidth]{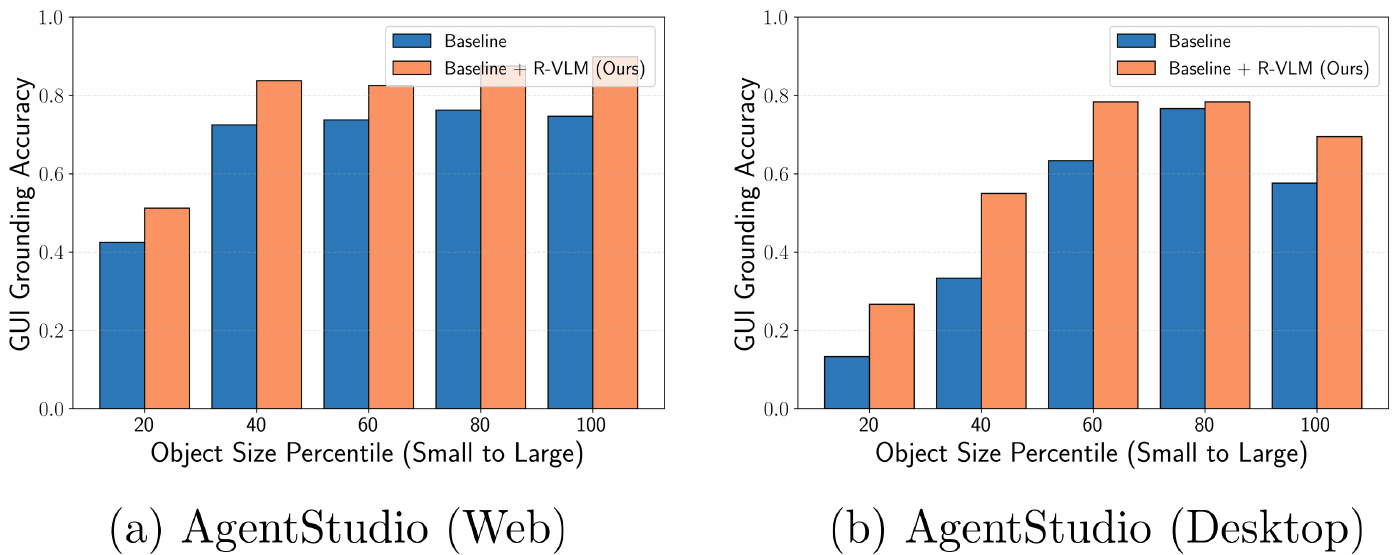}
  \caption{Grounding accuracy of GUI elements on the GroundUI-1K dataset from the AgentStudio benchmark, shown by object size percentiles (small to large) on x-axis. The blue bars and orange bars represent the grounding accuracy of baseline and our approach respectively.} 
  \label{fig:problem_2}
  \vspace{-6mm}
\end{figure}

From another perspective, we also observe the correlation between GUI object size and grounding accuracy of vision-language models (Figure~\ref{fig:problem_2}). We evaluate the visual agent model on the GroundUI-1K dataset from AgentStudio~\cite{zheng2024agentstudio}, which includes a wide range of object sizes as it combines data from multiple benchmarks. In Figure~\ref{fig:problem_2}, the x-axis represents the object size percentile, from small to large, showing that grounding accuracy is relatively low for smaller elements (blue bars). This outcome is somewhat expected, as accurately localizing small elements in high-resolution web or desktop screens with cluttered layouts poses a challenge. Motivated by these insights, we introduce tailored solutions to effectively improve the grounding accuracy of vision-language models in the following sections.

\subsection{Two-Stage Zoom-In Grounding}
\label{subsec:zoomingrounding}

For higher accuracy in locating GUI elements, we extract coordinates from zoomed-in region proposals that are likely to contain the target instance. The intuition behind here is that, given GUI environments often contain multiple elements with varying sizes, this process allows model to focus solely on object detection tasks within a small region, thereby significantly reducing the need to handle irrelevant context. This approach aligns with the conventional two-stage object detection strategies \cite{rcnn,fasterrcnn}, which separate the detection process into two steps, using the second step as a refinement stage to improve object regression accuracy. 

Our method begins with an initial prediction of a bounding box coordinate from the model during inference. We then crop a region centered around this initial prediction for zooming, assuming this region as a region proposal. Here the zoom-in scale is determined based on the estimated size of the GUI element, \ie, we zoom-in more for smaller elements, which visual agents often struggle to localize as discussed in Section~\ref{subsec:prelim}. Specifically, the zoom-in scale is set in proportion to the width and height of the initial prediction, and the cropped region is obtained by magnifying the initial prediction by a factor of \textit{k}.
The zoomed-in view around the initial prediction is then passed through the agent model, which predicts the coordinate within this selected region. Final coordinates are obtained by inverting the zoom-in view coordinate back to the original image perspective. It is worth noting that this two-stage verification approach significantly enhances localization accuracy in a \textit{post-hoc manner} without requiring additional instruction tuning. 

\paragraph{Instruction tuning on zoomed-in data}
Although our two-stage zoom-in grounding at inference works well without additional instruction tuning, we find that fine-tuning the model with zoom-in data and paired instructions simulating the zoom-in grounding process improves the model's precision in localizing coordinates within zoomed-in view of region proposals. We propose a simple yet effective zoom-in data generation pipeline derived from existing GUI grounding data, without introducing new data types. This process begins by determining the zoom-in region; rather than using ground-truth bounding boxes, we account for potential noise in initial predictions at the first inference stage. To simulate plausible noisy predictions, we generate bounding boxes by perturbing the ground-truth, ensuring a Generalized Intersection-over-Union (GIoU) score~\cite{giou} that exceeds a threshold $\sigma$. We adopt the GIoU metric here since it captures proximity to indicate whether regions are near or separate. With each generated bounding box, we crop and zoom-in around this area, pairing it with an instruction, such as ``\textit{Given the zoomed-in view centered on the initial prediction, predict a detailed bounding box for [INSTRUCTION]}". The label coordinates for these instructions are assigned by updating the ground-truth coordinates relative to the zoomed-in image. Empirical results confirm that training with this zoom-in region proposal and updated instruction format enhances grounding during the two-stage verification at inference. 

\subsection{IoU-Aware Weighted Cross-Entropy}
\label{subsec:iouloss}

A remaining challenge is that current models lack a learning signal indicating that their predictions should have a high IoU with the ground-truth bounding boxes. Existing visual agent models are trained by fitting only specific tokens from language-formatted coordinate labels using cross-entropy loss, resulting in the low IoU prediction patterns observed in Section~\ref{subsec:iouloss}. This raises a research question: \textit{how can we train a vision-language model to optimize for an IoU-like objective while it outputs numeric values as a sequence of tokens?} 
\begin{figure}[t]
  \centering
  \includegraphics[width=\linewidth]{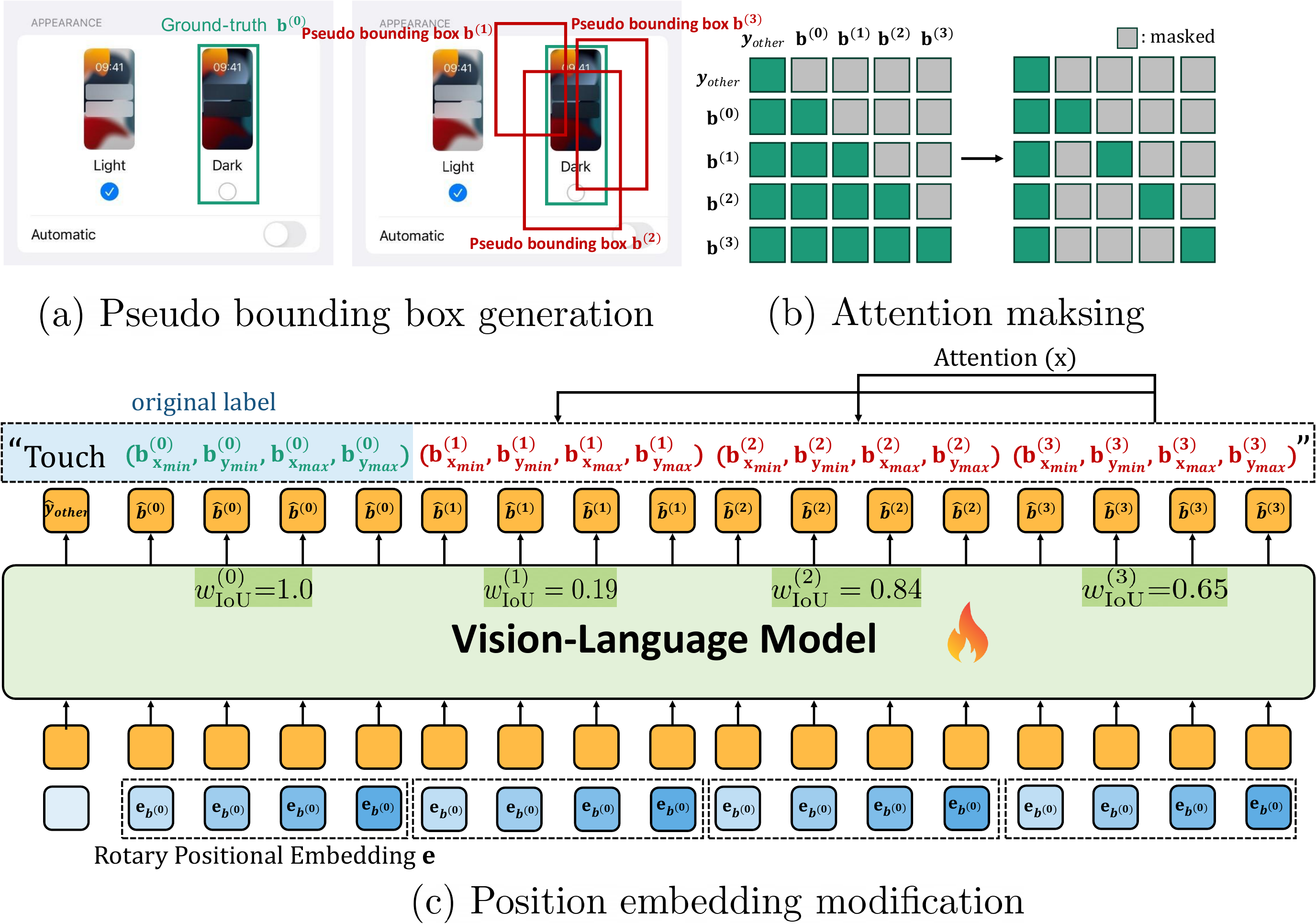}
   \caption{Efficient IoU-aware weighted cross-entropy computation. (a) $M$ pseudo boxes are generated from the ground-truth box using a GIoU threshold (here, $M$=3), and concatenated with the original label, enabling a single forward pass for $M$+1 predictions. (b) The attention map is masked to prevent pseudo boxes from attending to each other. (c) The positional embedding of the ground-truth box is assigned to all pseudo boxes, ensuring a single prediction at inference. Cross-entropy is weighted by GIoU relative to ground-truth.}
  \label{fig:iou_loss}
  \vspace{-0.6cm}
\end{figure}
With this goal in mind, we devise an IoU-aware weighted cross-entropy loss function. One straightforward approach for this loss is to generate pseudo bounding boxes, slightly deviated from ground-truth bounding boxes, to use as augmented labels.
Each pseudo bounding box is weighted in the cross-entropy loss calculation based on its IoU with the ground-truth: lower IoU values receive lower weights.  
While this simple design could guide the model toward better IoU alignment, it requires repetitive forward passes for the same GUI screenshot and user instruction (\eg, adding five different pseudo boxes for one label would require five forward passes). To solve this issue, we concatenate $M$ different pseudo bounding boxes with the label of ground-truth bounding box in a single forward pass (Figure~\ref{fig:iou_loss}), assigning each pseudo bounding box a different weight when computing the cross-entropy loss. This approach largely reduces computational costs by avoiding repetitive processing of the same GUI screenshot, though it results in the model predicting multiple boxes during inference. Thus, we introduce two simple modifications to ensure the model predicts only a single coordinate: \textbf{(1)} Modifying the attention mask to prevent attending to previous bounding box coordinates during training. \textbf{(2)} Adjusting the rotary positional embedding (RoPE)~\cite{rope} by assigning the positional embeddings of the first bounding box (\ie, ground-truth bounding box) to those of the pseudo bounding boxes.
By applying these modifications in the decoding steps, we achieve the effect of training the model on ($M$+1) distinct labels in a cost-efficient manner.

\noindent\textbf{Formulation} Given randomly generated M pseudo bounding boxes, denoted as $\mathbf{b}^{(1)}, ..., \mathbf{b}^{(M)}$, and a ground-truth bounding box $\mathbf{b}^{(0)}=(x_{\text{min}},y_{\text{min}},x_{\text{max}},y_{\text{max}})$, the label would be $\mathbf{y}=\{ y_{\text{other}},\mathbf{b}^{(0)},\mathbf{b}^{(1)}, ..., \mathbf{b}^{(M)}\}$ where $y_{\text{other}}$ represents other non-coordinate parts of the label (such as ``click") and consist of $N$ tokens. For simplicity, if we assume $\mathbf{b}^{(i)}$ as a single token (though it consists of multiple tokens in real), our IoU-aware weighted cross-entropy $\mathcal{L}^{\text{CE}}_{\text{IoU}}$ can be formulated as:
\begin{equation}
\resizebox{\columnwidth}{!}{
$\begin{aligned}
\mathcal{L}^{\text{CE}}_{\text{IoU}} = -\sum_{i=1}^{M}&w_{\text{IoU}}^{(i)}\mathbf{b}^{(i)}\log\hat{\mathbf{b}}^{(i)} - \sum_{j=1}^{N}y_{\text{other}}^{(j)}\log\hat{y}_{\text{other}}^{(j)}, \\
w^{(i)}_{\text{IoU}} &= 1+\frac{1}{2}\log(GIoU(\mathbf{b}^{(i)},\mathbf{b}^{(0)})),
\end{aligned}$
}
\end{equation}
where $GIoU(\mathbf{b}^{(i)},\mathbf{b}^{(0)})$ represents the generalized IoU between pseudo bounding box $\mathbf{b}^{(i)}$ and ground-truth bounding box $\mathbf{b}^{(0)}$. We use a log-scale of IoU to penalize low IoU pseudo boxes more while assigning weights close to 1 for high IoU pseudo boxes similar to the ground-truth. Here the pseudo bounding boxes are randomly generated to ensure a generalized IoU above a threshold, acquiring moderately deviated boxes from ground-truth boxes.

\section{Experiments}
\label{sec:experiments}

We conduct experiments on two different task settings: GUI grounding task and GUI agent task. Whereas the GUI grounding task focuses on evaluating the model's performance on localizing GUI elements, the GUI agent task focuses on the model's ability to navigate real-world GUI environments. We then verify our approach can indeed improve the IoU of predictions and grounding accuracy on small GUI elements. Furthermore, we show the benefit of two-stage zoom-in technique across multiple datasets in a training-free regime, demonstrating the plug-and-play nature of the proposed approach. Finally, we present detailed ablation studies to show how the proposed components drive the improvement in the grounding task.


\vspace{-0.2cm}
\subsection{Experiment Setup}
\vspace{-0.1cm}

\paragraph{Pretraining data}
We pretrain our model on the GUI grounding pretraining data from SeeClick~\cite{seeclick}, which comprises a collection of pretraining tasks designed for GUI grounding and understanding, such as predicting bounding box coordinates from user instructions, interpreting instructions from point coordinates, and GUI summarization. The dataset includes three distinct GUI domains - web, mobile, and desktop applications - and incorporates general vision-language paired instruction data from LLaVA-150k~\cite{liu2024llava} to preserve the model's reasoning and comprehension abilities across both visual and text inputs. This combined dataset results in a total of 1M samples with task-specific prompts.

\paragraph{Model architecture}
We apply R-VLM to the state-of-the-art approach SeeClick \cite{seeclick}.
Specifically, we use Qwen-VL 9.6B~\cite{qwenvl} as our backbone architecture. 
For training Qwen-VL, we use LoRA~\cite{hulora} and unfreeze the vision encoder path, facilitating the model to adapt toward GUI environments. Our pretraining is conducted on 8 NVIDIA A100 GPUs, and finetuning on agent tasks uses two A100 GPUs. 

\begin{table}[t]
\caption{Grounding results for GUI elements (text and icon) on the ScreenSpot benchmark. Grounding accuracy (click accuracy) is reported our method, along with both general-purpose large vision language models (upper) and vision-specialized agent models (lower).}
\vspace{-0.1in}
\begin{center}
\begin{small}
\setlength{\columnsep}{1.1pt}%
\renewcommand{\arraystretch}{1.0}
\begin{adjustbox}{width=0.99\linewidth}
\begin{tabular}{l cc cc cc c }
\toprule
\multirow{2}{*}{\textbf{Method}} 
&\multicolumn{2}{c}{\texttt{Mobile}} &   \multicolumn{2}{c}{\texttt{Desktop}} &  \multicolumn{2}{c}{\texttt{Web}} & \multirow{2}{*}{Average} \\
\cline{2-3} \cline{4-5}\cline{6-7}
& Text & Icon & Text & Icon & Text & Icon & \\
\cline{1-8}
                GPT-4V &  22.6 & 24.5 & 20.2 & 11.8 & 9.2 & 8.8 & 16.2 \\
\hdashline
                Fuyu & 41.0 & 1.3 & 33.0 & 3.6 & 33.9 & 4.4 & 19.5 \\
                CogAgent & 67.0 & 24.0 & 74.2 & 20.0 & 70.4 & 28.6 & 47.4 \\
                SeeClick & 78.0 & 52.0 & 72.2 & 30.0 & 55.7 & 32.5 & 53.4  \\
                \hline                    
                \rowcolor{lightgray}$\text{\textbf{R-VLM}}$ & \textbf{85.0} & \textbf{61.1} &\textbf{81.4} & \textbf{52.8} & \textbf{66.5} & \textbf{51.4} &\textbf{66.3 \textcolor{forestgreen}{(+12.9)}}  \\
\bottomrule
\end{tabular}
\end{adjustbox}
\end{small}
\end{center}
\label{tb:main_screenspot}
\vspace{-0.1in}
\end{table}

\begin{table}[t]
\caption{GUI grounding results on the GroundUI-1K dataset from the AgentStudio benchmark, combining MoTiF and OmniAct with screenshots sourced from other agent tasks. Grounding accuracy is reported for our approach and other vision language models.}
\vspace{-0.1in}
\begin{center}
\setlength{\columnsep}{1pt}%
\begin{adjustbox}{width=0.99\linewidth}
\begin{tabular}{l cccc}
\toprule
\multirow{2}{*}{\textbf{Method}} &  \multicolumn{3}{c}{GroundUI-1K (AgentStudio)} &  \multirow{2}{*}{Average} \\
\cline{2-4}
&\texttt{Web} & \texttt{Desktop} & \texttt{Mobile} & \\
\cline{1-5}

                    CogVLM2-19B & 2.5 & 2.7 & 5.3 & 3.4 \\ 
                    GPT-4o & 7.5 & 8.3 & 26.3 & 13.4 \\
                    Claude-3.5-Sonnet & 13.0 & 14.0 & 26.3 & 17.3 \\
                    CogAgent  & 25.3 & 15.7 & 35.7 & 25.5 \\       
                    Gemini-1.5-pro & 31.2 & 24.3 & 51.3 & 35.2 \\
                    SeeClick & 64.3 & 44.3 & 73.7 & 61.1 \\
                    \hline                    
                    \rowcolor{lightgray}$\text{\textbf{R-VLM}}$ & \textbf{76.5} & \textbf{65.3} & \textbf{79.7} & \textbf{74.1 \textcolor{forestgreen}{(+13.0)}}\\
\bottomrule
\end{tabular}
\end{adjustbox}
\end{center}
\label{tb:main_agentstudio}
  \vspace{-0.24in}
\end{table}

\begin{table*}[t]
\caption{Action matching score and click accuracy (grounding accuracy for click actions) of the proposed R-VLM compared to baselines on the AITW mobile automation tasks benchmark.}
\vspace{-0.1in}
\begin{center}
\setlength{\columnsep}{1pt}%
\begin{adjustbox}{width=0.75\linewidth}
\begin{tabular}{l cccccccc}
\toprule
\multirow{2}{*}{\textbf{Method}} &  \multicolumn{5}{c}{Android In The Wild (AITW)} &  \multirow{2}{*}{Average} & \multirow{2}{*}{Click Acc.} \\
\cline{2-6}
&\texttt{General} & \texttt{Install} & \texttt{GoogleApps} & \texttt{Single} & \texttt{WebShopping} & \\
\cline{1-8}
                ChatGPT-CoT & 5.9 & 4.4 & 10.5 & 9.4 & 8.4 & 7.7 & - \\
                PaLM2-CoT & - & - & - & - & - & 39.6 & - \\
                GPT-4V & 41.7 & 42.6 & 49.8 & 72.8 & 45.7 & 50.5 & - \\
                Qwen-VL & 49.5 & 59.9 & 46.9 & 64.7 & 50.7 & 54.3 & 57.4 \\
                SeeClick & 54.0 & 66.4 & 54.9 & 63.5 & 57.6 & 59.3 & 66.4 \\
                \hline            
                \rowcolor{lightgray}$\text{\textbf{R-VLM}}$ & \textbf{59.9} & \textbf{70.6} &	\textbf{59.6} & 72.5 & \textbf{61.7} & \textbf{64.9  \textcolor{forestgreen}{(+5.6)}} & \textbf{71.0 \textcolor{forestgreen}{(+4.6)}} \\

\bottomrule
\end{tabular}
\end{adjustbox}
\end{center}
\label{tb:main_aitw}
  \vspace{-3mm}
\end{table*}
\subsection{GUI Grounding Task}
\label{subsec:guigrounding}

\paragraph{Evaluation details} First, we evaluate our pretrained R-VLM on GUI grounding tasks using two benchmarks - ScreenSpot~\cite{seeclick} and GroundUI-1K~\cite{zheng2024agentstudio}, which consists of direct grounding instructions (\eg ``Click to add title") paired with GUI screenshots and corresponding bounding box coordinates. For GroundUI-1K, screenshots are sourced from the various existing grounding benchmarks~\cite{omniact,mind2web,motif} across web, desktop applications, and mobile platforms, as well as from agent task datasets. Likewise, ScreenSpot samples include mobile, desktop (OS and applications), and web platforms. This dataset comprises over 1,200 instructions of real-world user interfaces, each associated with executable GUI icons and text buttons.

\paragraph{Results}
In Table~\ref{tb:main_screenspot}, we report the grounding accuracy of R-VLM and baselines, including general-purpose VLM GPT-4V~\cite{achiam2023gpt}, and GUI-specific visual agents - Fuyu-8B~\cite{fuyu-8b}, CogAgent-18B~\cite{cogagent}, SeeClick-9.6B~\cite{seeclick} - on the ScreenSpot benchmark. Our framework significantly enhances grounding accuracy by an average of 12.9\% over SeeClick, despite identical pretraining data and model architecture, with the only difference being the inclusion of zoomed-in data. Notably, we observe that accuracy improvements for icon grounding in desktop and web environments are particularly large, where small icons tend to be cluttered. These results suggest that our two-stage zoom-in grounding and IoU-aware loss effectively enable precise coordinate localization, even for tightly spaced GUI elements. We also evaluate our approach on the GroundUI-1K benchmark, which has broader data distribution, in comparison to recent large VLMs. As shown in Table~\ref{tb:main_agentstudio}, our approach achieves the highest grounding accuracy across all GUI platforms, with a  13\% improvement. 

\begin{table*}[h]
\caption{Web navigation results on the Mind2Web benchmark without HTML documents (vision-only). Element grounding accuracy (Ele. Acc.), operation F1 score (Op. F1), and step success rate (Step SR) are reported for our method and baseline models.}
\vspace{-0.1in}
\begin{center}
\setlength{\columnsep}{0.8pt}%
\begin{adjustbox}{width=0.75\linewidth}
\begin{tabular}{l ccccccccc}
\toprule
& \multicolumn{9}{c}{Mind2Web (\textit{vision-only})} \\
\multirow{1}{*}{\textbf{Method}} &  \multicolumn{3}{c}{\texttt{Cross-Task}} &  \multicolumn{3}{c}{\texttt{Cross-Website}}
&  \multicolumn{3}{c}{\texttt{Cross-Domain}}\\
\cline{2-4} \cline{5-7} \cline{8-10}
& \begin{small}Ele. Acc.\end{small} & \begin{small}Op. F1\end{small} & \begin{small}Step SR\end{small} & \begin{small}Ele. Acc.\end{small} & \begin{small}Op. F1\end{small} & \begin{small}Step SR\end{small} & \begin{small}Ele. Acc.\end{small} & \begin{small}Op. F1\end{small} & \begin{small}Step SR\end{small} \\
\cline{1-10}

                    Qwen-VL & 15.9 & 86.7 & 13.3 & 13.2 & 83.5 & 9.2 & 14.1 & 84.3 & 12.0 \\ 
                    SeeClick & 28.3 & 87.0 & 25.5 & 21.4 & 80.6 & 16.4 & 23.2 & 84.8 & 20.8 \\
                    \hline                    
                    \rowcolor{lightgray}$\text{\textbf{R-VLM}}$ &\textbf{ 31.6} & \textbf{88.0}& \textbf{28.7 \textcolor{forestgreen}{(+3.2)}} &\textbf{29.5} &\textbf{ 84.9} & \textbf{26.1 \textcolor{forestgreen}{(+9.7)}} & \textbf{26.7} & \textbf{85.3} & \textbf{24.3 \textcolor{forestgreen}{(+3.5)}} \\
\bottomrule
\end{tabular}
\end{adjustbox}
\end{center}
\label{tb:main_mind2web}
\vspace{-0.4cm}
\end{table*}

\subsection{GUI Navigation Tasks}
\label{subsec:agent_tasks}

\paragraph{Evaluation details} We conduct experiments on downstream navigation tasks to show that improved accuracy in GUI grounding can boost the success rate of navigation tasks. Unlike direct GUI grounding, GUI navigation requires the model to predict the next action and coordinates based on high-level instructions and a sequence of history actions. For navigation tasks, we ensure that the zoomed-in region includes all areas of history actions, and also update the coordinates of history actions in the prompt according to the zoomed-in region during second-stage grounding. We fine-tune our pretrained model for each task: Android In The Wild (AITW)~\cite{rawles2024androidinthewild} and Mind2Web~\cite{mind2web}. The AITW is a mobile automation dataset containing 30k task instructions and corresponding sequential human actions. We follow the dataset split of \citet{seeclick} to evaluate models on unseen instructions and avoid overfitting as the split in previous studies includes test set instructions during training. Mind2Web is a web navigation dataset with 2k human action trajectories navigating real-world websites. 
We use GUI screenshots as input and acquire target bounding boxes from the raw data of Mind2Web, consistent with previous visual agent studies. 


\begin{figure}[t]
  \centering
  \includegraphics[width=\linewidth]{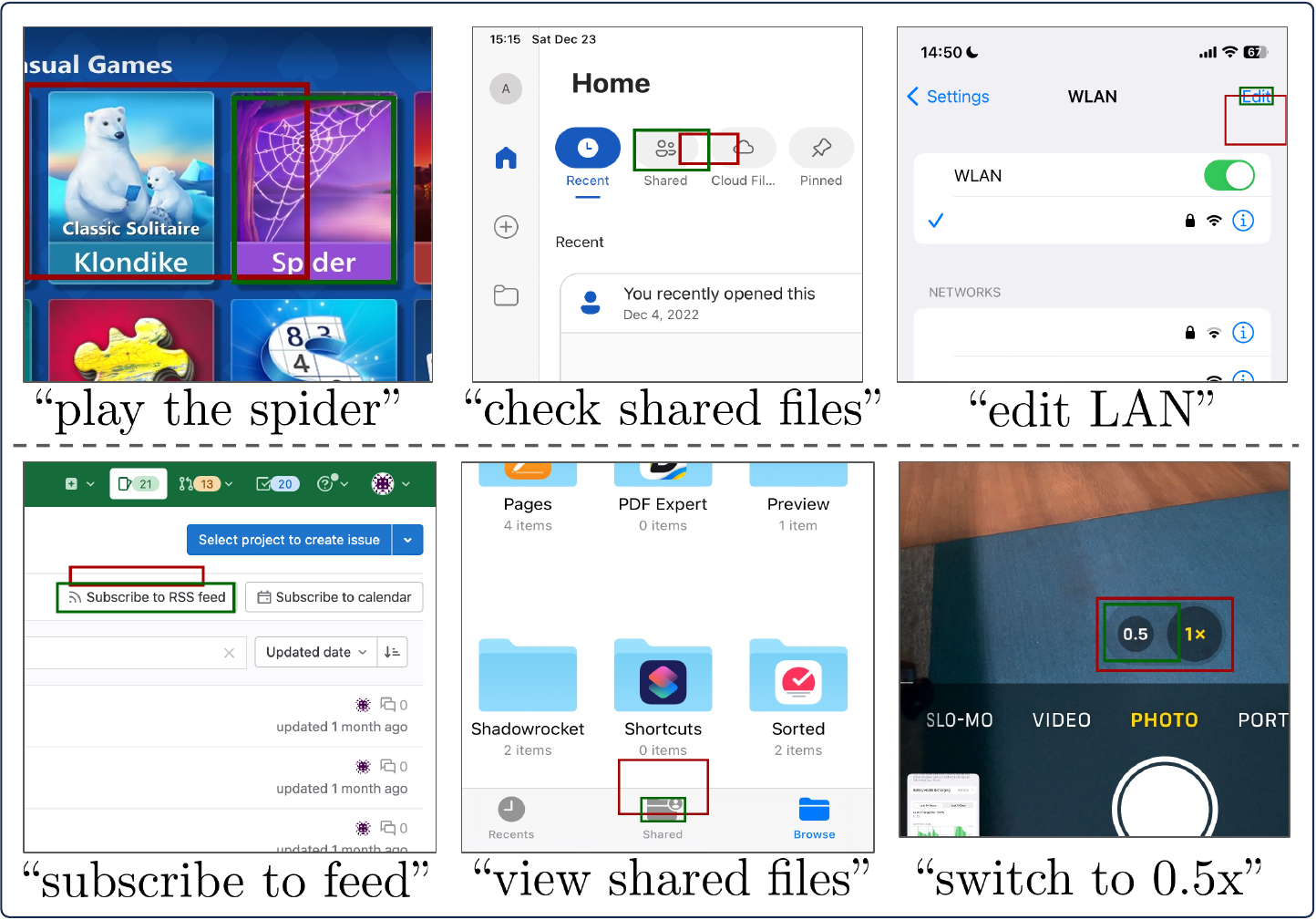}
  \caption{Qualitative examples of GUI grounding. Predictions from our R-VLM are shown in \textcolor{forestgreen}{green}, while the baseline predictions are shown in \red{red}.} 
  \label{fig:examples}
  \vspace{-0.2in}
\end{figure}%
\paragraph{Results} For AITW dataset, screen-wise action matching scores and averaged click accuracy are reported across 5 different mobile environments (Table~\ref{tb:main_aitw}). We compare our method with recent vision language models and also with API-based large language models. As shown in Table~\ref{tb:main_aitw}, R-VLM achieves the highest grounding accuracy, with an average improvement of 5.6\%, resulting in consistently high action matching scores. For the Mind2Web dataset (Table~\ref{tb:main_mind2web}), we benchmark our framework against other visual agent baselines using visual inputs and report element grounding accuracy, operation F1 score, and step success rate. R-VLM improves accuracy by an average of 4.9\% over the baseline, leading to a higher step success rate. These results imply that visual agents armed with our strategy can achieve high success rate in automating human action trajectories.

\subsection{Further Analysis}
\label{subsec:further_analysis_main}
We probe that R-VLM effectively addresses the challenges in GUI grounding as discussed in Section~\ref{subsec:prelim}. Here, we compare our approach to the SeeClick baseline on the same GUI grounding dataset GroundUI-1K. We focus on two aspects: 1) IoU-value trends in grounding predictions and 2) the impact of GUI element size. In Figure~\ref{fig:examples} and Appendix \ref{sec:appx_qualiative_analysis}, we also present qualitative examples of GUI grounding for the baseline~\cite{seeclick} and R-VLM, demonstrating that R-VLM achieves precise localization.

\begin{figure}[t]
  \centering
  \includegraphics[width=\linewidth]{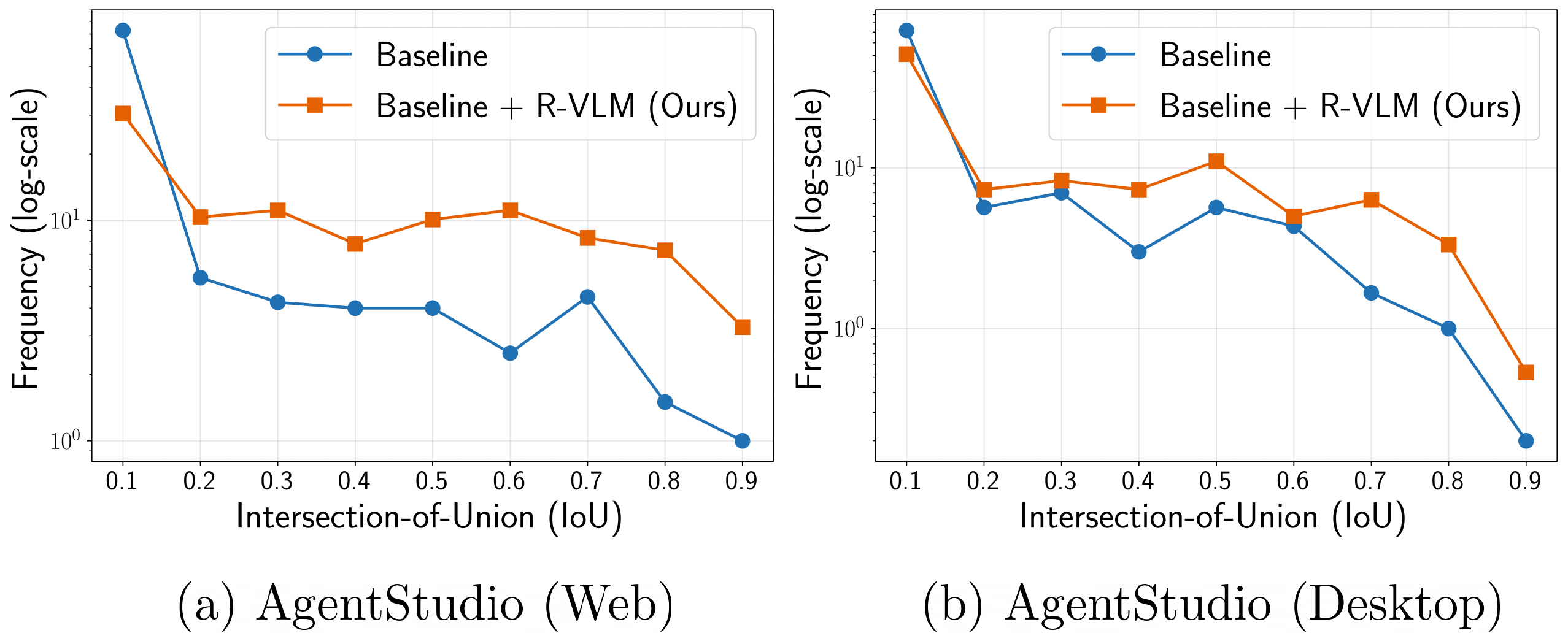}
  \vspace{-0.2in}
  \caption{Comparison of IoU histograms for GUI elements grounding predictions between our R-VLM and the baseline on the GroundUI-1K benchmark from AgentStudio. Our approach exhibits a higher frequency of high IoU-value prediction patterns. }
  \label{fig:iou_comparison}
\vspace{-0.18in}
\end{figure}

\paragraph{IoU distribution in GUI grounding}
Figure~\ref{fig:iou_comparison} presents the IoU value distribution of GUI grounding predictions across web and desktop environments. R-VLM demonstrates a notable shift toward higher IoU values, reflecting more precise localization of GUI elements. This finding suggests that the proposed IoU-aware weighted loss effectively guides the model toward high-IoU predictions, reducing overfitting to specific numerical tokens.

\paragraph{Object sizes} We investigate how grounding accuracy varies with GUI element size after applying our method (Figure~\ref{fig:problem_2}). While the baseline struggles with localizing smaller elements, R-VLM markedly boosts grounding accuracy for these cases. By leveraging fine-grained coordinate regression from the zoomed-in region proposal, R-VLM achieves precise localization, even for small instances, facilitating action execution against these elements.

\begin{table*}[t]
\caption{Results on two benchmarks comparing baselines and our two-stage zoom-in grounding, applied at inference in a training-free setting. 
Left: grounding accuracy for GUI elements (text and icon) on the ScreenSpot benchmark. 
Right: element accuracy on the Multimodal-Mind2Web benchmark with planner-generated grounding queries.}
\vspace{-0.1in}
\begin{center}
\begin{small}
\renewcommand{\arraystretch}{1.1}
\begin{adjustbox}{width=0.85\linewidth}
\begin{tabular}{lc|cc cc cc c|ccc}
\toprule
\multirow{3}{*}{\textbf{Method}} & \multirow{3}{*}{\begin{footnotesize}\shortstack[1]{Training \\Dataset}\end{footnotesize}} & \multicolumn{7}{c|}{ScreenSpot} & \multicolumn{3}{c}{Multimodal-Mind2Web} \\
 &  & \multicolumn{2}{c}{\texttt{Mobile}} & \multicolumn{2}{c}{\texttt{Desktop}} & \multicolumn{2}{c}{\texttt{Web}} & \multirow{2}{*}{Avg.} & \multirow{2}{*}{\begin{footnotesize}\texttt{Cross-Task}\end{footnotesize}} & \multirow{2}{*}{\begin{footnotesize}\texttt{Cross-Website}\end{footnotesize}} & \multirow{2}{*}{\begin{footnotesize}\texttt{Cross-Domain}\end{footnotesize}} \\
 \cline{3-4}\cline{5-6}\cline{7-8}
 && Text & Icon & Text & Icon & Text & Icon & \\
 \cline{1-12}
UGround & \begin{footnotesize}10M\end{footnotesize} & 82.8 & 60.3 & 82.5 & 63.6 & 80.4 & 70.4 & 73.3 & 47.7 & 46.0 & 46.3 \\
\rowcolor{lightgray}
$\text{\textbf{R-VLM}}_{\text{UGround}}$ & \begin{footnotesize}10M\end{footnotesize} & 79.9 & \textbf{65.1} & \textbf{85.1} & \textbf{66.4} & \textbf{80.9} & \textbf{73.3} & \textbf{75.1} & \textbf{51.4 \textcolor{forestgreen}{(+3.7)}} & \textbf{49.2 \textcolor{forestgreen}{(+3.2)}} & \textbf{48.9 \textcolor{forestgreen}{(+2.6)}} \\
\bottomrule
\end{tabular}
\end{adjustbox}
\end{small}
\end{center}
\vspace{-0.15in}
\label{tb:merged_trainfree}
\end{table*}

\subsection{Training-Free Zoom-In Grounding}

We proceed by validating that our two-stage zoom-in grounding in inference can be seamlessly integrated into pretrained VLMs in a \textit{training-free} manner (\ie, without instruction tuning on zoomed-in data or training with IoU-aware loss). To show the versatility of this method, we conduct experiments in settings different from our main experiments. We choose UGround~\cite{uground}, which is built on a different VLM architecture LLaVA-NeXT~\cite{liu2024llava}, as our baseline model. UGround has been pretrained on 10 millions of high-resolution GUI using over 100 H100 GPUs and is much stronger than SeeClick \cite{seeclick}. We test our training-free zoom-grounding in two tasks: GUI grounding setting on ScreenSpot, and the agent task guided by a \textit{large language model planner} on Multimodal-Mind2Web~\cite{multimoda-mind2web}. 
In this setting, GUI agents operate in coordination with a planning model, prompting the model to perform grounding based on queries generated by the planner.

As presented in Table~\ref{tb:merged_trainfree} (left), incorporating our zoom-in grounding approach post hoc significantly improves GUI localization accuracy for UGround. In particular, even for a large-scale GUI pretrained model, our approach enhances icon/widget localization accuracy by an average of 3.5\%. For the web agent task (Table~\ref{tb:merged_trainfree} (right)), where grounding is guided by planner-generated queries, our method consistently exhibits superior element accuracy over all settings, with an average increase of 3.2\%. These results indicate that our approach provides robust, training-free improvements in grounding accuracy for large-scale agents. 

\vspace{-0.05in}
\subsection{Ablation Study}
\vspace{-0.01in}
\paragraph{Step-wise evaluation}
In Table~\ref{tb:ablation_study}, we present a stepwise evaluation to assess the contribution of each module in R-VLM: 
(1) employing two-stage zoom-in grounding at inference, (2) incorporating instruction tuning with zoomed-in data, and (3) combining IoU-aware loss with zoom-in grounding. The results clearly demonstrate that each component contributes to more precise GUI localization. Interestingly, IoU-aware loss and zoom-in grounding exhibit complementary synergy, as IoU-aware loss drives the model to predict more accurate region proposal in the first stage.

\paragraph{The number of zoom-in steps}
To examine whether additional zoom-in stages can further improve grounding accuracy, we conduct experiments by varying the number of zooming stages from 2 to 4, while also measuring the corresponding inference latency. These experiments are performed on the ScreenSpot benchmark using Qwen-VL-9.6B~\cite{qwenvl} trained with our method. Notably, only the number of zoom-in stages at inference time is adjusted, and the inference latency is measured using a single RTX 3090 GPU. As shown in Table~\ref{tb:zoom_in_steps}, increasing the number of zoom-in stages from 2 to 4 improves the grounding accuracy by 1.1\%, at the cost of 2x inference latency. These results indicate that the two-stage zoom-in approach provides the best trade-off between grounding accuracy and inference latency, while increasing the number of zoom-in stages may be a viable option for performance-sensitive applications.

\begin{table}[t]
\caption{Effect of increasing zoom-in steps from 2 to 4 on GUI grounding accuracy and inference latency (measured in seconds per sample) on the ScreenSpot benchmark.}
\vspace{-0.13in}
\begin{center}
\begin{small}
\setlength{\columnsep}{1.0pt}%
\begin{adjustbox}{width=1.\linewidth}
\renewcommand{\arraystretch}{1.1}
\begin{tabular}{lc|cc cc cc c}
\toprule
 \multirow{2}{*}{\textbf{Method}} & \begin{footnotesize}\multirow{2}{*}{\shortstack[1]{\textbf{Zoom-in} \\\textbf{Steps}}}\end{footnotesize} & \multicolumn{2}{c}{\texttt{Mobile}} &   \multicolumn{2}{c}{\texttt{Desktop}} &  \multicolumn{2}{c}{\texttt{Web}} & \multirow{2}{*}{\shortstack[1]{Inference \\Latency}} \\
\cline{3-4}\cline{5-6}\cline{7-8}
&& Text & Icon & Text & Icon & Text & Icon & \\  
\cline{1-9}

SeeClick &\begin{footnotesize}-\end{footnotesize} & 78.0 & 52.0 & 72.2 & 30.0 & 55.7 & 32.5 &  $1.4\,\mathrm{s}/\mathrm{sample}$  \\
\cdashline{1-9}
\rowcolor{lightgray} & \begin{footnotesize} $\times 2$ \end{footnotesize} & 85.0 & 61.1 &\textbf{ 81.4} & 52.8 & 66.5 & 51.4 & $2.7\,\mathrm{s}/\mathrm{sample}$\\

\rowcolor{lightgray}$\text{\textbf{R-VLM}}$& \begin{footnotesize}$\times 3$ \end{footnotesize} & 84.6  & \textbf{64.6} & 80.4 & 51.4 & 66.1 & \textbf{53.4} & $4.1\,\mathrm{s}/\mathrm{sample}$\\

\rowcolor{lightgray} & \begin{footnotesize}$\times 4$ \end{footnotesize} & \textbf{86.4}  & 62.9 & 80.9 & \textbf{54.2} & \textbf{68.7} & 51.5 & $5.6\,\mathrm{s}/\mathrm{sample}$\\

\bottomrule
\end{tabular}
\end{adjustbox}
\end{small}
\end{center}
\label{tb:zoom_in_steps}
\vspace{-0.05in}
\end{table}

\begin{table}[t]
\caption{Ablation study of R-VLM on three different modules in ScreenSpot benchmark.}
\vspace{-0.1in}
\begin{center}
\setlength{\columnsep}{0.8pt}%
\begin{adjustbox}{width=1.\linewidth}
\begin{tabular}{ccc|cc cc cc c }
\toprule
\begin{footnotesize}\multirow{2}{*}{\shortstack[1]{\textbf{IoU-aware} \\ \textbf{Weighted CE}}}\end{footnotesize} & \begin{footnotesize}\multirow{2}{*}{\shortstack[1]{\textbf{Zoom-in} \\ \textbf{Inst. tuning}}}\end{footnotesize} & \begin{footnotesize}\multirow{2}{*}{\shortstack[1]{\textbf{Zoom-in} \\ \textbf{Inference}}}\end{footnotesize} & \multicolumn{2}{c}{\texttt{Mobile}} &   \multicolumn{2}{c}{\texttt{Desktop}} &  \multicolumn{2}{c}{\texttt{Web}} & \multirow{2}{*}{Average} \\
\cline{4-5} \cline{6-7}\cline{8-9}
&&& Text & Icon & Text & Icon & Text & Icon & \\
\cline{1-10}
                &&& 78.0 & 52.0 & 72.2 & 30.0 & 55.7 & 32.5 & 53.4  \\
                && \cmark & 79.1 & 60.3 & 77.8 & 47.1 & 63.9 & 42.2 & 61.7 \textcolor{forestgreen}{(+8.3)}  \\
                
                &\cmark & \cmark & 82.1 & 61.6 & 74.7 & 51.6 & 65.1 & 48.2 & 63.9  \textcolor{forestgreen}{(+10.5)} \\
                
                \cmark & \cmark & \cmark & \textbf{85.0} & 61.1 &\textbf{ 81.4} & \textbf{52.8} & \textbf{66.5} & \textbf{51.4} &\textbf{66.4 \textcolor{forestgreen}{(+12.9)}}  \\
\bottomrule
\end{tabular}
\end{adjustbox}
\end{center}
\label{tb:ablation_study}
\vspace{-0.4cm}
\end{table}



\section{Conclusion}
We propose R-VLM, a novel region-aware vision language model designed for precise GUI grounding.
R-VLM incorporates a region proposal mechanism and an IoU-aware objective function, inspired by conventional object detection approaches. 
Specifically, R-VLM uses initial predictions as region proposals to derive more fine-grained predictions from zoomed-in proposals.
The IoU-aware objective function provides a learning signal that guides predictions toward high IoU values relative to ground truth locations in a cost-efficient manner.
Elaborate experiments show that our approach significantly improves grounding accuracy across diverse GUI environments, accurately localizing even small GUI elements.

\section*{Limitations}
Although our approach shows promising accuracy on GUI grounding and GUI navigation tasks, the accuracy of our approach is upper-bounded by the recall of the first-stage prediction - if the first-stage prediction largely deviates from the target element, i.e., zoom-in region not containing the target element, our approach will not be able to recover from the error. We provide an analysis of how this issue affects the failure cases of our approach in Appendix~\ref{subsec:appx_failure_case_analysis}. In the future, we will explore generating multiple bounding boxes as region proposals to improve the recall of the first-stage prediction to further improve the accuracy of GUI grounding.

\bibliography{custom}
\clearpage
\appendix
In the Appendix, we present additional experimental results to provide an in-depth analysis of our proposed R-VLM in Section~\ref{sec:appx_exp_results}. Section~\ref{sec:exp_setup} describes the implementation details, and Section~\ref{sec:appx_qualiative_analysis} provides GUI grounding examples of R-VLM. We also include the pseudocode for R-VLM in Section~\ref{sec:appx_pseudo_code} and discuss related work from another domain in Section~\ref{sec:appx_related_work_vqa}.

\section{Additional Experimental Results}
\label{sec:appx_exp_results}

We first demonstrate that our method can be seamlessly extended to \textbf{point coordinate predictions}. Next, we investigate the effect of \textbf{varying the number of pseudo bounding boxes}, analyze the \textbf{computation cost} of our IoU-aware weighted cross-entropy loss, and examine how the \textbf{zoom-in scale} impacts grounding accuracy. We also conduct an ablation study on the GroundUI-1K dataset. 
\begin{table}[t]
\caption{Comparison of action matching score and click accuracy on the Android In The Wild dataset between the proposed R-VLM applied to point prediction and bounding box prediction.}
\begin{center}
\setlength{\columnsep}{0.7pt}%
\begin{adjustbox}{width=1.\linewidth}
\begin{tabular}{l cccccccc}
\toprule
\multirow{2}{*}{\textbf{Method}} &  \multicolumn{5}{c}{Android In The Wild (AITW)} &  \multirow{2}{*}{Avg.} & \multirow{2}{*}{Click Acc.} \\
\cline{2-6}
& \begin{footnotesize}\texttt{General}\end{footnotesize} &  \begin{footnotesize}\texttt{Install}\end{footnotesize} &  \begin{footnotesize}\texttt{GoogleApps}\end{footnotesize} &  \begin{footnotesize}\texttt{Single}\end{footnotesize} &  \begin{footnotesize}\texttt{WebShopping}\end{footnotesize} & \\
\cline{1-8}
                SeeClick & 54.0 & 66.4 & 54.9 & 63.5 & 57.6 & 59.3 & 66.4 \\
                 \rowcolor{lightgray}\textbf{R-VLM} (Point Pred.) & \textbf{61.4 }& 66.8 & 57.3 & 72.1 & \textbf{66.6} & \textbf{64.8} & \textbf{70.5 \textcolor{forestgreen}{(+4.1)}} \\
                 \rowcolor{lightgray}\textbf{R-VLM} (Box Pred.) & 59.9 & \textbf{70.6} &	\textbf{59.6} & \textbf{72.5} & 61.7 & \textbf{64.9} & \textbf{71.0 \textcolor{forestgreen}{(+4.6)}} \\

\bottomrule
\end{tabular}
\end{adjustbox}
\end{center}
\label{tb:appx_point_aitw}
\vspace{-0.1in}
\end{table} 
\subsection{Extension to Point Coordinate Prediction}
\label{subsec:appx_pointpred}

Our two key components can be applied for point coordinate prediction tasks in grounding GUI elements. For the two-stage zoom-in grounding, after obtaining the initial ``point coordinate'' prediction, we define a region proposal as a rectangular area centered on the initial point prediction, maintaining the aspect ratio of the original image. This region is then zoomed in to allow the model to refine the initially predicted point coordinates at the second stage grounding. For the IoU-aware weighted cross-entropy loss, the metric can be modified from IoU to Euclidean distance for point prediction tasks while preserving the core concepts. After randomly generating pseudo points slightly deviated from the ground-truth point, the weights of these pseudo ``point" labels are determined based on their Euclidean distance from ground-truths.

To verify our approach on the point coordinate prediction task, we conduct two experiments: (1) On the Android In The Wild (AITW) benchmark~\cite{rawles2024androidinthewild}, we measure grounding accuracy and action-matching score using the model pretrained with the Euclidean distance-aware weighted cross-entropy loss. Inference is conducted using region proposals centered on the initial point coordinate predictions. (2) we measure grounding accuracy on the ScreenSpot dataset~\cite{seeclick} using two-stage zoom-in grounding at inference only (training-free). For both experiments, the width and height of region proposals are set to 0.3x of the original image resolution.

\begin{table}[t]
\caption{Grounding accuracy for GUI elements on the ScreenSpot benchmark, comparing bounding box prediction and point prediction with the two-stage zoom-in grounding applied at inference only (training-free setting).}
\begin{center}
\begin{small}
\setlength{\columnsep}{1.0pt}%
\begin{adjustbox}{width=1.\linewidth}
\begin{tabular}{l|cc cc cc c}
\toprule
 \multirow{2}{*}{\shortstack[l]{\textbf{Method} \\(\textit{Training-free})}} & \multicolumn{2}{c}{\texttt{Mobile}} &   \multicolumn{2}{c}{\texttt{Desktop}} &  \multicolumn{2}{c}{\texttt{Web}} & \multirow{2}{*}{Avg.} \\
\cline{2-3}\cline{4-5}\cline{6-7}
& Text & Icon & Text & Icon & Text & Icon & \\  
\cline{1-8}
SeeClick & 78.0 & 52.0 & 72.2 & 30.0 & 55.7 & 32.5 & 53.4    \\

  \rowcolor{lightgray}\textbf{R-VLM} (Point Pred.) &  \textbf{79.9}& 54.2 & 73.7 & \textbf{52.1} & 61.7 & \textbf{49.0} &\textbf{61.8}  \\
  
   \rowcolor{lightgray}\textbf{R-VLM} (Box Pred.) & 79.1 & \textbf{60.3} & \textbf{77.8} & 47.1 & \textbf{63.9} & 42.2 & \textbf{61.7} \\
\bottomrule
\end{tabular}
\end{adjustbox}
\end{small}
\end{center}
\label{tb:appx_point_screenspot}
\end{table}

\begin{figure}[t]
  \centering
  \includegraphics[width=\linewidth]{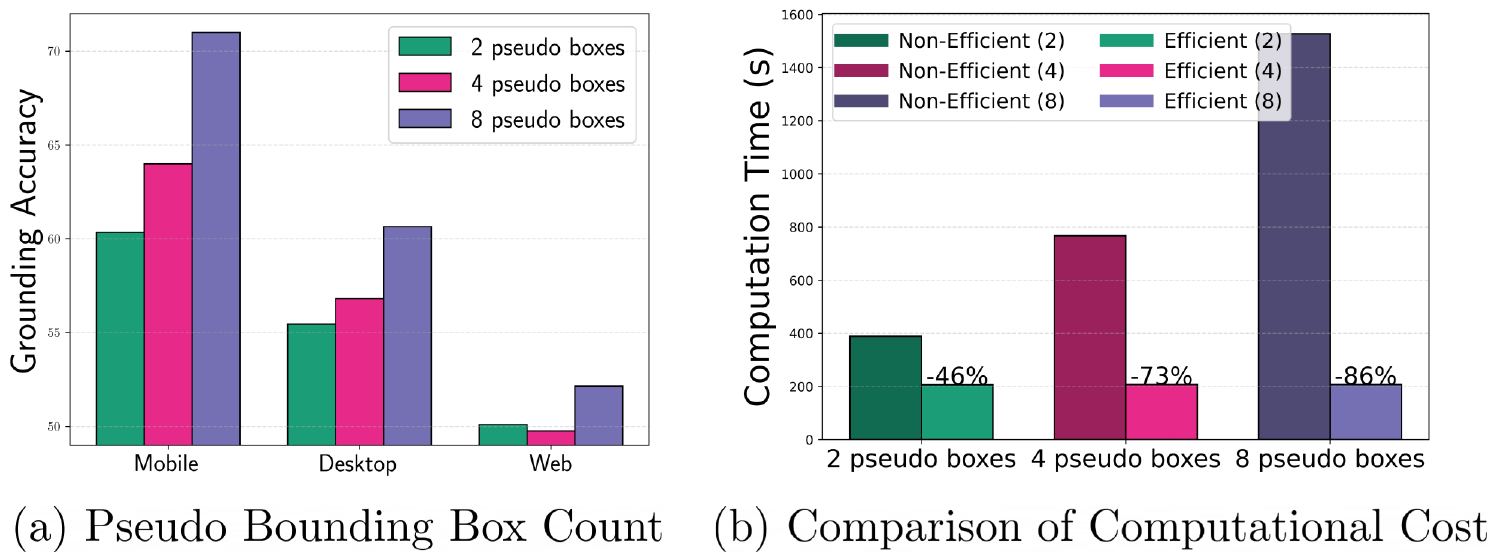}
  \caption{Results of the ablation study on IoU-aware weighted cross-entropy on the ScreenSpot. (a) Grounding accuracies when varying the number of pseudo bounding boxes used in the IoU-aware loss. (b) Computation cost comparison between our cost-efficient IoU-aware loss and the standard IoU-aware weighted loss that treats pseudo bounding boxes as independent instructions.}
  \label{fig:appx_iouloss_ablation}
\end{figure}%

As confirmed in Table~\ref{tb:appx_point_aitw} and \ref{tb:appx_point_screenspot}, our approach, when adapted to point predictions, achieves grounding accuracies comparable to - and in some cases exceeding - those of bounding box predictions.

\subsection{Ablation Study on IoU-aware Weighted CE}
\label{subsec:appx_bbox_ablation}

In this section, we delve deeper into the analysis of our proposed IoU-aware weighted cross-entropy loss from two perspectives. First, we investigate whether increasing the number of pseudo bounding boxes leads to further improvements in grounding accuracy. Second, as discussed in Section~\ref{subsec:iouloss}, we introduce two practical techniques to reduce the computational cost of using IoU-aware loss with pseudo bounding box labels. To evaluate the effectiveness of these strategies, we compare our approach to a counterpart that adjusts loss weights based on the IoU of pseudo bounding boxes but treats them as independent instructions, without applying our cost-saving techniques.

\paragraph{Effect of pseudo bounding box count}
We evaluate the model on the ScreenSpot benchmark by varying the number of pseudo bounding boxes used during pretraining with the IoU-aware loss. As shown in Figure~\ref{fig:appx_iouloss_ablation} (a), the results indicate that increasing the number of pseudo bounding box labels from 2 to 8 consistently enhances grounding accuracy across diverse GUI environments. This suggests that the model benefits from learning with pseudo bounding boxes with a broader range of IoU values and corresponding loss weights. 

\begin{table}[t]
\caption{Web navigation results of R-VLM at three different zoom-in levels - High, Mid, and Low- for two-stage zoom-in grounding on the Mind2Web (vision-only). Element grounding accuracy (Ele. Acc.) and step success rate (Step SR) are reported.}
\begin{center}
\setlength{\columnsep}{0.8pt}%
\begin{adjustbox}{width=1.\linewidth}
\begin{tabular}{l cccccc}
\toprule
& \multicolumn{6}{c}{Mind2Web (\textit{vision-only})} \\
\multirow{1}{*}{\textbf{Method}} &  \multicolumn{2}{c}{\texttt{Cross-Task}} &  \multicolumn{2}{c}{\texttt{Cross-Website}}
&  \multicolumn{2}{c}{\texttt{Cross-Domain}}\\
\cline{2-3} \cline{4-5} \cline{6-7}
& \begin{small}Ele. Acc.\end{small} & \begin{small}Step SR\end{small} & \begin{small}Ele. Acc.\end{small} & \begin{small}Step SR\end{small} & \begin{small}Ele. Acc.\end{small} & \begin{small}Step SR\end{small} \\
\cline{1-7}

                    Qwen-VL & 15.9 & 13.3 & 13.2 & 9.2 & 14.1 & 12.0 \\ 
                    SeeClick & 28.3 & 25.5 & 21.4 & 16.4 & 23.2 & 20.8 \\
                    \hline                    
                    \rowcolor{lightgray} \textbf{R-VLM} \begin{footnotesize}(High-zoom)\end{footnotesize}  & 31.2 & 28.1 \textcolor{forestgreen}{(+2.6)} & 27.2 & 24.5 \textcolor{forestgreen}{(+8.1)} & \textbf{27.2} & \textbf{24.5 \textcolor{forestgreen}{(+3.7)}} \\
                    \rowcolor{lightgray} \textbf{R-VLM} \begin{footnotesize}(Mid-zoom)\end{footnotesize}  &\textbf{31.6} & \textbf{28.7 \textcolor{forestgreen}{(+3.2)}} &\textbf{29.5} & \textbf{26.1 \textcolor{forestgreen}{(+9.7)}} & 26.7 & 24.3 \textcolor{forestgreen}{(+3.5)} \\
                    \rowcolor{lightgray} \textbf{R-VLM} \begin{footnotesize}(Low-zoom)\end{footnotesize}  & 30.8 & 27.9 \textcolor{forestgreen}{(+2.4)} & 27.3 & 22.7 \textcolor{forestgreen}{(+6.3)} & 25.6 & 22.3 \textcolor{forestgreen}{(+1.5)} \\
\bottomrule
\end{tabular}
\end{adjustbox}
\end{center}
\label{tb:appx_mind2web_zoominscale}
\end{table}

\paragraph{Training cost comparison}
To assess the practicality of our cost-efficient IoU-aware loss, we compare its training cost to a baseline implementation that treats each pseudo bounding box as an independent instruction. Figure~\ref{fig:appx_iouloss_ablation} (b) reports the average training time for 200 instructions, conducted under identical conditions using an NVIDIA A6000 GPU. Remarkably, our approach reduces training time by 73\% and 86\% for 4 and 8 pseudo bounding boxes, respectively, while maintaining consistent computational efficiency over different numbers of pseudo labels. These results show that our method provides IoU-guided learning signals effectively without introducing computational overhead, making it a practical choice for large-scale training.

\subsection{Impact of Zoom-in Scale Factor}
\label{subsect:appx_zoomin}

To explore the impact and sensitivity of zoom-in scale on two-stage zoom-in grounding, we conduct experiments by varying the zoom-in scale across three different levels - High, Mid, and Low. For same GUI target element, we set region proposals by scaling the width and height of initial prediction by factors of 3x (High-zoom), 5x (Mid-zoom), and 7x. These region proposals are then cropped and zoomed-in to the original image resolution for second-stage grounding. As presented in Table~\ref{tb:appx_mind2web_zoominscale}, our method consistently improves element grounding accuracy, leading to increases in step success rates across all zoom-in scales. Overall, High-zoom and Mid-zoom exhibit superior localization accuracy compared to the Low-zoom.

\subsection{Ablation Study on GroundUI-1K}
\label{subsec:appx_groundui}
In addition to the stepwise evaluation of key modules of R-VLM in the main paper, we present ablation study results on an additional GUI grounding dataset, GroundUI-1k dataset~\cite{zheng2024agentstudio} (Table~\ref{tb:appx_ablation_study_groundui}). These results further demonstrate the contribution of each module toward precise GUI grounding, highlighting the complementary synergy between zoom-in grounding and the IoU-aware weighted cross-entropy loss.
\begin{table}[t]
\caption{Ablation study of R-VLM on three different modules in GroundUI-1K benchmark.}
\begin{center}
\setlength{\columnsep}{1pt}%
\begin{adjustbox}{width=1.\linewidth}
\begin{tabular}{ccc|c c c c }
\toprule
\begin{footnotesize}\multirow{2}{*}{\shortstack[1]{\textbf{IoU-aware} \\ \textbf{Weighted CE}}}\end{footnotesize} & \begin{footnotesize}\multirow{2}{*}{\shortstack[1]{\textbf{Zoom-in} \\ \textbf{Inst. tuning}}}\end{footnotesize} & \begin{footnotesize}\multirow{2}{*}{\shortstack[1]{\textbf{Zoom-in} \\ \textbf{Inference}}}\end{footnotesize} &  \multicolumn{4}{c}{GroundUI-1K (AgentStudio)}  \\
\cline{4-7}
& & & \multicolumn{1}{c}{\texttt{Mobile}} &   \multicolumn{1}{c}{\texttt{Desktop}} &  \multicolumn{1}{c}{\texttt{Web}} & Avg. \\
\cline{1-7}
                &&&  64.3 & 44.3 & 73.7 & 61.1   \\
                && \cmark & 74.3 & 62.7 & 76.0 & 71.2 \textcolor{forestgreen}{(+10.1)}  \\
                
                &\cmark & \cmark & 74.5 & 63.3 & 76.0 & 71.6 \textcolor{forestgreen}{(+10.5)}\\
                
                \cmark & \cmark & \cmark & \textbf{76.5} & \textbf{65.3} & \textbf{79.7} & \textbf{74.1 \textcolor{forestgreen}{(+13.0)}}  \\
\bottomrule
\end{tabular}
\end{adjustbox}
\end{center}
\label{tb:appx_ablation_study_groundui}
\end{table}

\begin{figure*}[t]
  \centering
  \includegraphics[width=\linewidth]{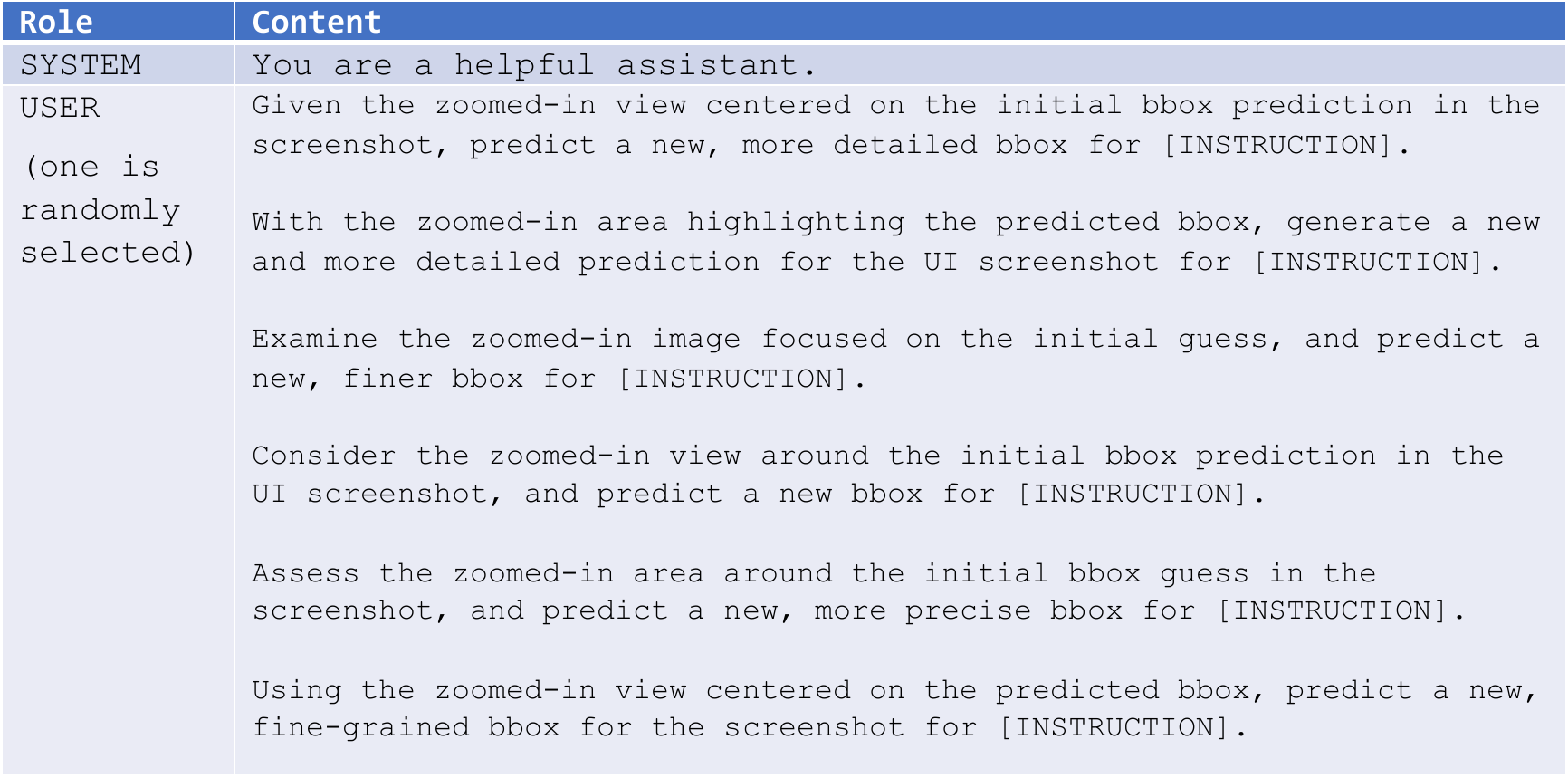}
  \caption{Examples of a set of prompts for instruction tuning on zoomed-in data.}
  \label{fig:appx_zoomin_instructions}
  \vspace{-0.1in}
\end{figure*}%

\begin{figure*}[t]
  \centering
  \includegraphics[width=\linewidth]{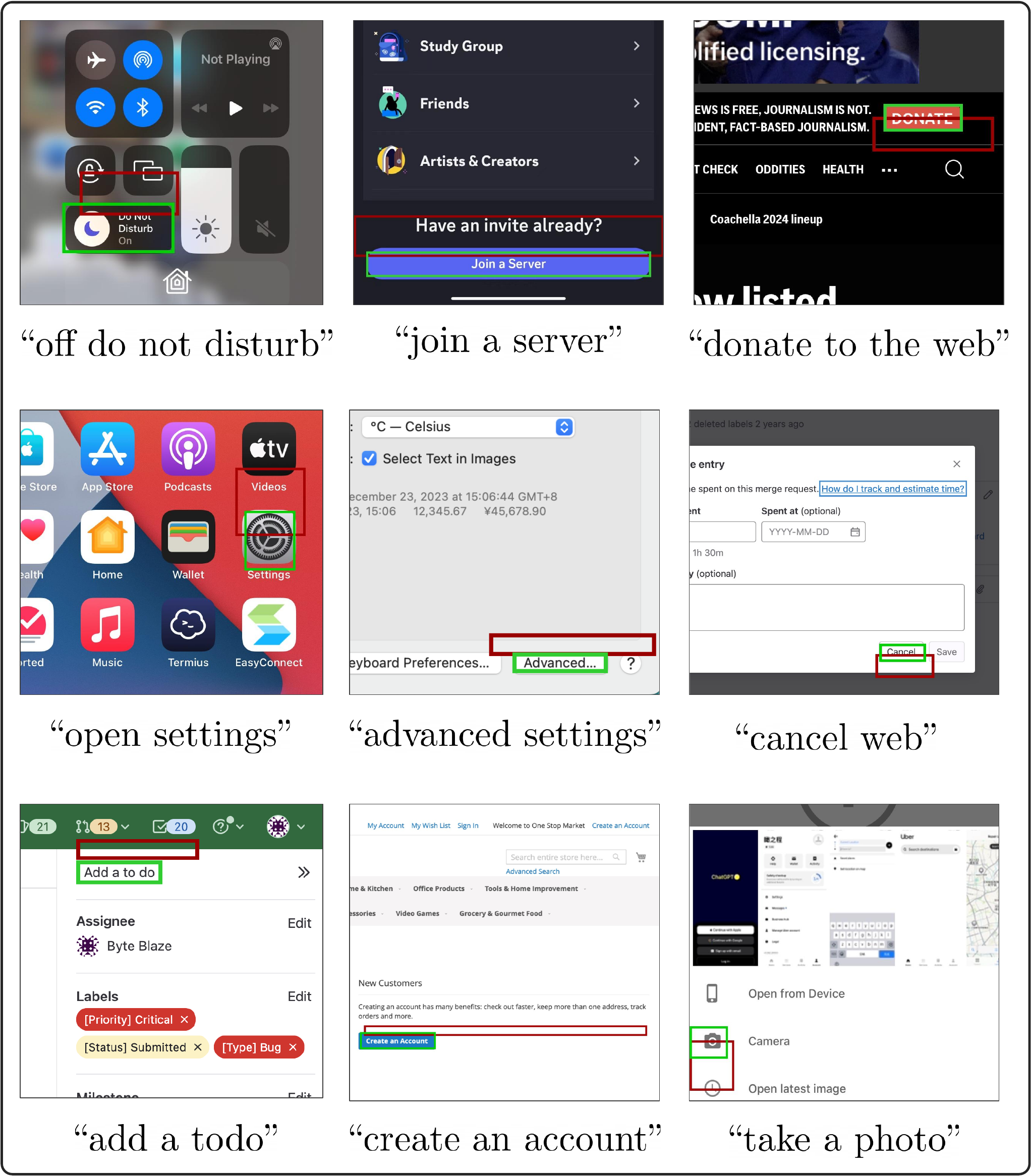}
  \caption{Qualitative examples of GUI grounding. Predictions from our R-VLM are shown in \textcolor{forestgreen}{\textbf{green}}, while the baseline predictions are shown in \red{\textbf{red}}. Due to the two-stage zoom-in strategy and IoU aware training R-VLM can accurately ground challenging GUI elements.}
  \label{fig:appx_qualitative_1}
\end{figure*}%

\subsection{Impact of Initial Region Proposals on Grounding Failures}
\label{subsec:appx_failure_case_analysis}
The accuracy of the R-VLM method is limited by the recall rate of the first-stage prediction as mentioned in the Limitations section. Specifically, if the first-stage prediction deviates too far from the target element, resulting in the zoomed-in region not containing the target element, R-VLM will not be able to recover from this error. To better understand how this issue affects the failure cases of our approach, we conduct an analysis of the grounding failure cases (\ie, instances where the center point of the predicted box does not fall within the ground-truth region, leading to action failure). Specifically, we examine the probability that the ground-truth region is included in the initial region proposal among failure cases on the ScreenSpot benchmark, using Qwen-VL-9.6B~\cite{qwenvl} trained with our method.
\begin{table}[h]
\caption{Probability that the ground-truth region of the grounding failure cases is included in the initial region proposal.}
\vspace{-0.1in}
\begin{center}
\begin{small}
\setlength{\columnsep}{0.9pt}%
\begin{adjustbox}{width=0.85\linewidth}
\begin{tabular}{l|cccc}
\toprule
\multirow{2}{*}{\textbf{Element Type}} & \multicolumn{4}{c}{ScreenSpot} \\
& \begin{footnotesize}\texttt{Mobile}\end{footnotesize} &  \begin{footnotesize}\texttt{Desktop}\end{footnotesize} & \begin{footnotesize}\texttt{Web}\end{footnotesize} & \begin{footnotesize}\texttt{Avg.}\end{footnotesize}   \\ 
\cline{1-5}
Text & 81.0 & 80.6 & 65.2 & 75.6 \\
Icon/Widget & 59.6 & 61.7 & 62.7 & 61.3 \\

\bottomrule
\end{tabular}
\end{adjustbox}
\end{small}
\end{center}
\label{tb:initial_recall_rate}
\vspace{-0.2in}
\end{table}

As shown in Table~\ref{tb:initial_recall_rate}, more than half of the current failure cases include the ground-truth region in the initial region proposal, implying that while the model often captures a nearby area, it fails to localize precisely. These results suggest that R-VLM has the room to rectify a substantial proportion of failure cases - more than half at most. However, our approach is still upper-bounded by the recall rate of the first-stage prediction. In future work, we will explore how to improve the recall rate of the first-stage prediction.



\section{Detailed Experiment Setup}
\label{sec:exp_setup}

\subsection{Implementation Details}

For implementing our approach for experiments, we adopt a zoom-in scaling factor $k$ from \{5, 7\}, and the number of pseudo bounding boxes for IoU-aware loss is set to 4. The pseudo bounding boxes are generated using Generalized Intersection-over-Union (GIoU)~\cite{giou} values greater than 0.3. For GUI agent tasks, which incorporate the coordinates of previous actions in prompts during multi-step instructions, we update the coordinates of previous actions relative to the zoomed-in region proposals, ensuring that the region proposals encompass the entire area of the previous actions. For instruction tuning on zoomed-in data, we use a set of prompts tailored to zoomed-in data, with examples provided in Figure~\ref{fig:appx_zoomin_instructions}. 
\vspace{-0.01in}
\subsection{Training Settings}
\vspace{-0.02in}
We pretrain Qwen-VL~\cite{qwenvl} on GUI grounding data from \cite{seeclick}. LoRA~\cite{hulora} is adopted to update the model while the Vision Transformer (ViT) encoder path is unfrozen to adapt its parameters to GUI screenshots. The model is trained on 8 NVIDIA A100 GPUs with a batch size of 64, a weight decay of 0.1, and for 3 epochs. The AdamW optimizer~\cite{adamw} is used with a learning rate of 3e-5 and $\beta_{2}$ set to 0.95. In addition to the existing GUI pretraining data, we include 600K zoomed-in data samples centered on randomly generated bounding boxes, which have a GIoU greater than -0.2 with ground-truth bounding boxes. For finetuning on GUI agent tasks, we maintain the same training settings as in the pretraining process, except for training the model for 10 epochs, following the setup in SeeClick \cite{seeclick}.

\vspace{-0.05in}
\section{Qualitative Case Study: R-VLM}
\label{sec:appx_qualiative_analysis}
\begin{figure*}[t]
  \centering
  \includegraphics[width=\linewidth]{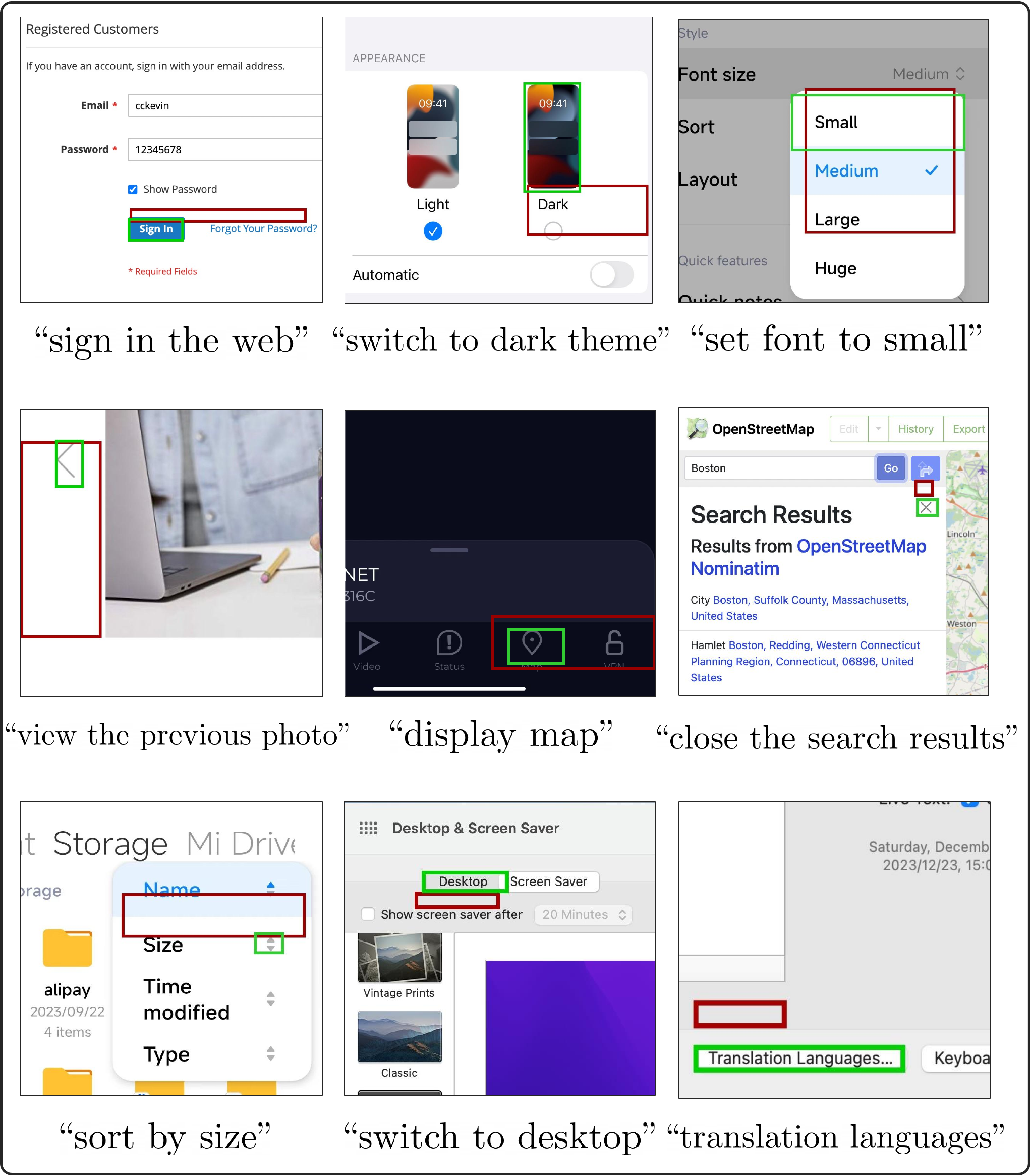}
  \caption{Qualitative examples of GUI grounding. Predictions from our R-VLM are shown in \textcolor{forestgreen}{\textbf{green}}, while the baseline predictions are shown in \red{\textbf{red}}. Due to the two-stage zoom-in strategy and IoU aware training R-VLM can accurately ground challenging GUI elements.}
  \label{fig:avppx_qualitative_2}
\end{figure*}

\vspace{-0.05in}
We conduct a case study comparing R-VLM to the SeeClick \cite{seeclick} baseline (using the identical model and pretraining data). Figure~\ref{fig:appx_qualitative_1} and Figure~\ref{fig:avppx_qualitative_2} illustrate examples of grounding predictions made by R-VLM, contrasted with the baseline, which fails to execute the actions. Predictions by R-VLM are shown in green, while those by the baseline are depicted in red. R-VLM achieves precise grounding on the GUI element by excluding unrelated regions and accurately identifying the target layout, whereas the baseline often captures nearby but incorrect regions.

\clearpage
\section{PyTorch Pseudocode}
\label{sec:appx_pseudo_code}
\begin{lstlisting}[language=Python, caption=Determining region proposal via zoom-in on initial prediction., label={lst:zoom_crop}]
def zoom_in_region_proposal(image: Tensor, bbox: List[float]) -> Tuple[Tensor, Tuple[int, int, int, int]]:
    """
    Args:
        image: Tensor of shape (3, H, W)
        bbox: Predicted box in [xmin, ymin, xmax, ymax] format

    Returns:
        zoomed_image: Zoom-in region centered on the initial predicted bounding box
        crop_coords: Crop coordinates in original image space, used to map prediction back from the zoomed view
    """
    _, H, W = image.shape

    # Convert predicted bbox to absolute coordinates
    xmin, ymin = bbox[0] * W, bbox[1] * H
    xmax, ymax = bbox[2] * W, bbox[3] * H

    # Compute center and size of the predicted box
    xc = (xmin + xmax) / 2
    yc = (ymin + ymax) / 2
    bw, bh = xmax - xmin, ymax - ymin

    # Expand region proposal (e.g., 5x enlarged around the predicted bbox)
    cw, ch = 5 * bw, 5 * bh

    # Clamp the zoom-in crop region to image bounds
    xmin_c = max(int(xc - cw / 2), 0)
    ymin_c = max(int(yc - ch / 2), 0)
    xmax_c = min(int(xc + cw / 2), W)
    ymax_c = min(int(yc + ch / 2), H)

    # Crop the region
    cropped = image[:, ymin_c:ymax_c, xmin_c:xmax_c]

    # Resize cropped region back to original resolution
    zoomed_image = F.resize(cropped, size=[H, W])

    return zoomed_image, (xmin_c, ymin_c, xmax_c, ymax_c)
\end{lstlisting}

\begin{lstlisting}[language=Python, 
  caption=Generating pseudo bounding boxes for IoU-aware weighted cross-entropy loss., 
  label={lst:giou_bboxes}, 
  basicstyle=\ttfamily\scriptsize]
def generate_giou_bboxes(gt_box: Tensor, n_outputs: int, num_candidates: int = 100, threshold: float = 0.3):
    """
    Perturbs a ground-truth box to generate pseudo boxes for IoU-weighted CE loss.

    Args:
        gt_box: (4,) in [xmin, ymin, xmax, ymax], normalized
        n_outputs: number of outputs; threshold: minimum GIoU
    Returns:
        valid_boxes (n_outputs, 4); valid_gious (n_outputs,)
    """
    
    shifts = torch.rand(num_candidates, 4) * 4 - 2.0
    scales = torch.rand(num_candidates, 2) * 0.4 + 0.8

    w, h = gt_box[2] - gt_box[0], gt_box[3] - gt_box[1]
    boxes = gt_box.unsqueeze(0).repeat(num_candidates, 1)
    boxes[:, [0, 2]] += shifts[:, [0, 2]] * w
    boxes[:, [1, 3]] += shifts[:, [1, 3]] * h

    cx = (boxes[:, 0] + boxes[:, 2]) / 2; cy = (boxes[:, 1] + boxes[:, 3]) / 2
    nw = w * scales[:, 0]; nh = h * scales[:, 1]
    boxes[:, 0] = cx - nw / 2; boxes[:, 2] = cx + nw / 2
    boxes[:, 1] = cy - nh / 2; boxes[:, 3] = cy + nh / 2

    boxes = torch.clamp(boxes, 0, 1)
    boxes = torch.round(boxes * 100) / 100

    gious = generalized_iou(gt_box.unsqueeze(0), boxes)
    mask = gious >= threshold
    return boxes[mask][:n_outputs], gious[mask][:n_outputs]
\end{lstlisting}
We provide PyTorch pseudocode snippets for two key functions: zoom-in procedure based on initial predicted box to obtain a region proposal (Code~\ref{lst:zoom_crop}), and pseudo bounding box generation used in IoU-aware weighted cross-entropy loss (Code~\ref{lst:giou_bboxes}).

\vspace{-0.05in}
\section{Related Work Beyond Our Task}
\label{sec:appx_related_work_vqa}
Recent studies in visual question answering also explore region-focused strategies such as cropping and zoom-in mechanisms~\citep{wu2024v,shen2024zoomeye}. While these approaches share high-level similarities with ours, they differ in both objectives and methodology, and our work introduces a distinct framework tailored to GUI grounding.

\citet{wu2024v} identifies visual instances in complex scenes by leveraging a large language model (LLM) to iteratively guide a search process based on common-sense reasoning (\eg, A is likely to be near B). This method relies on iterative cropping, external LLM inference, and finetuning. In contrast, our goal is to enhance the grounding capability of vision language models (VLMs) themselves, \ie, predicting high-IoU coordinates in natural language without auxiliary modules or finetuning. This distinction is critical for GUI automation, where inaccurate grounding directly impacts action execution. Rather than using an external LLM, our two-stage zoom-in grounding is a lightweight, self-improving strategy that enhances grounding accuracy using the model itself. We further introduce IoU-aware loss and zoom-in instruction tuning, designed to support high-IoU grounding within this two-stage framework. 

Regarding~\citet{shen2024zoomeye}, although it also employs zoom-in via tree-based image exploration, its goal aligns with \cite{wu2024v}: identifying instances for visual question answering. The method recursively splits images into quadrants and zooms into high-confidence regions. While effective for general visual search, this design is less applicable to GUI agent tasks due to several challenges: 1) GUI images are often long and high-resolution, making exhaustive tree traversal costly. 2) GUI agents typically require multi-step action sequences, where incorrect grounding can compound over time. 3) The domain shift between general vision tasks and GUI environments makes model confidence unreliable for recursive search. In summary, despite shared high-level intuition, our two-stage zoom-in grounding is uniquely suited to GUI agents, offering a practical and scalable solution for improving grounding accuracy in VLMs.

\end{document}